\documentclass[11pt]{article}
\pdfoutput=1
\usepackage{microtype}
\usepackage{graphicx}
\usepackage{subfigure} 
\usepackage{booktabs} 
\usepackage{natbib}
\usepackage{epsfig,epsf,fancybox}
\usepackage{amsmath}
\usepackage{mathrsfs}
\usepackage{amssymb}
\usepackage{color}
\usepackage{multirow}
\usepackage{paralist}
\usepackage{verbatim}
\usepackage{algorithm}
\usepackage{algorithmic}
\usepackage{galois}
\usepackage{boxedminipage}
\usepackage{accents}
\usepackage{stmaryrd}
\usepackage[bottom]{footmisc}
\usepackage{microtype}
\usepackage{booktabs} 
\usepackage{amsmath}
\usepackage{amssymb}
\usepackage{amsthm}
\usepackage[mathscr]{euscript}
\usepackage{nicefrac}
\usepackage{wrapfig}
\newcommand{\tablewhitespace}{\addlinespace[0.3em]}

\usepackage{natbib}
\usepackage{tcolorbox}
\usepackage{url}
\usepackage{listings}

\usepackage[colorlinks, linkcolor={blue!70!black}, citecolor={blue!70!black}, urlcolor={blue!70!black}]{hyperref}

\usepackage{xcolor}

\usepackage{algorithm}
\usepackage{algorithmic}

\usepackage{mathtools}
\usepackage[flushleft]{threeparttable}
\usepackage{varwidth}
\usepackage{xfp, siunitx}
\usepackage{tikz}
\usepackage{bm}
\usepackage{tcolorbox}
\usepackage[mathscr]{euscript}

\definecolor{tab10_blue}{rgb}{0.121, 0.466, 0.705}
\definecolor{tab10_orange}{rgb}{1.0,   0.498, 0.054}
\definecolor{tab10_green}{rgb}{0.172, 0.627, 0.172}
\definecolor{tab10_red}{rgb}{0.839, 0.152, 0.156}
\definecolor{tab10_purple}{rgb}{0.580, 0.403, 0.741}
\definecolor{tab10_brown}{rgb}{0.549, 0.337, 0.294}
\definecolor{tab10_pink}{rgb}{0.890, 0.466, 0.760}
\definecolor{tab10_gray}{rgb}{0.498, 0.498, 0.498}
\definecolor{tab10_olive}{rgb}{0.737, 0.741, 0.133}
\definecolor{tab10_cyan}{rgb}{0.090, 0.745, 0.811}

\usepackage{enumitem}

\newcommand{\speedup}[1]{{\color{gray}(\ifdim #1 pt > 0.3pt #1\else $< #1$\fi{}$\times$)}}
\newcommand{\scale}[2]{\fpeval{(#2 - #1) / 45}}
\newcommand{\barplot}[2]{ \tikz[baseline=-0.75ex]{
  \begin{scope}
    \clip (0,-6pt) rectangle (60pt, 6pt);
    \draw[black!5,line width=6pt] (0,0) -- (60pt,0);
    \draw[tab10_blue!80,line width=6pt] (0,0) -- ({\scale{#1}{#2} * 60pt},0);
  \end{scope}}
  }

\definecolor{Gray}{gray}{0.85}

\usepackage
[a4paper,
left=2.5cm,
right=2.5cm,
top=3cm,
bottom=4cm]
{geometry}

\begin{document}


\begin{center}

{\bf{\LARGE{{TCT}: Convexifying Federated Learning using  \\\vspace{2.5mm}  Bootstrapped Neural Tangent Kernels}}}

\vspace*{.25in}
{\large{
\begin{tabular}{c}
Yaodong Yu$^{\diamond}$ \quad Alexander Wei$^{\diamond}$ \quad Sai Praneeth Karimireddy$^{\diamond}$ 
\end{tabular}
}}

\vspace*{.05in}
{\large{
\begin{tabular}{c}
Yi Ma$^{\diamond}$ \quad
Michael I. Jordan$^{\diamond, \dagger}$\\
\end{tabular}
}}

\vspace*{.15in}
\begin{tabular}{c}
Department of Electrical Engineering and Computer Sciences$^\diamond$ \\
Department of Statistics$^\dagger$ \\ 
University of California, Berkeley
\end{tabular}

\vspace*{.1in}

\begin{abstract}
\noindent
State-of-the-art federated learning methods can perform far worse than their centralized counterparts when clients have dissimilar data distributions.
For neural networks, even when centralized SGD easily finds a solution that is simultaneously performant for all clients, current federated optimization methods fail to converge to a comparable solution. 
We show that this performance disparity can largely be attributed to optimization challenges presented by \emph{nonconvexity}. 
Specifically, we find that the early layers of the network do learn useful features, but the final layers fail to make use of them. That is, federated optimization applied to this non-convex problem distorts the learning of the final layers. 
Leveraging this observation, we propose a \emph{\textbf{T}rain-\textbf{C}onvexify-\textbf{T}rain}~(TCT) procedure to sidestep this issue: first, learn features using off-the-shelf methods (e.g., FedAvg); then, optimize a \emph{convexified} problem obtained from the network's empirical neural tangent kernel approximation. 
Our technique yields accuracy improvements of up to $+36\%$ on FMNIST and $+37\%$ on CIFAR10 when clients have dissimilar data. 
\end{abstract}

\end{center}

\vspace{-1em}
\section{Introduction}\label{sec:intro}
Federated learning is a newly emerging paradigm for machine learning where multiple data holders (clients) collaborate to train a model on their combined dataset. Clients only share partially trained models and other statistics computed from their dataset, keeping their raw data local and private~\citep{mcmahan2017communication,kairouz2019federated}. By obviating the need for a third party to collect and store clients' data, federated learning has several advantages over the classical, centralized paradigm~\citep{dean2012large,iandola2016firecaffe,goyal2017accurate}: it ensures clients' consent is tied to the specific task at hand by requiring active participation of the clients in training, confers some basic level of privacy, and has the potential to make machine learning more participatory in general~\citep{paml2020,jones2020nonrivalry}. Further, widespread legislation of data portability and privacy requirements (such as GDPR and CCPA) might even make federated learning a necessity~\citep{pentland2021building}.

Collaboration among clients is most attractive when clients have very different subsets of the combined dataset (\emph{data heterogeneity}). For example, different autonomous driving companies may only be able to collect data in weather conditions specific to their location, whereas their vehicles would need to function under all conditions. In such a scenario, it would be mutually beneficial for companies in geographically diverse locations to collaborate and share data with each other. Further, in such settings, clients are physically separated and connected by ad-hoc networks with large latencies and limited bandwidth. This is especially true when clients are edge devices such as mobile phones, IoT sensors, etc. Thus, \emph{communication efficiency} is crucial for practical federated learning. However, it is precisely under such circumstances (large data heterogeneity and low communication) that current algorithms fail dramatically~\cite[etc.]{hsieh2020non,li2020federated,karimireddy2020scaffold,reddi2021adaptive,wang2020tackling,acar2021federated,li2021model,afonin2021towards,wang2021field}. This motivates our central question:
{\emph{Why do current federated methods fail in the face of data heterogeneity---and how can we fix them?}}

\vspace{-0.15in}
\paragraph{Our solution.} We make two main observations: (i) We show that, even with data heterogeneity, linear models can be trained in a federated manner through gradient correction techniques such as SCAFFOLD~\citep{karimireddy2020scaffold}. While this observation is promising, it alone remains limited, as linear models are not rich enough to solve practical problems of interest (e.g., those that require feature learning). (ii)~We shed light on why current federated algorithms struggle to train deep, nonconvex models. We observe that the failure of existing methods for neural networks is not uniform across the layers. The early layers of the network do in fact learn useful features, but the final layers fail to make use of them. Specifically, federated optimization applied to this nonconvex problem results in distorted final layers.

These observations suggest a \emph{train-convexify-train} federated algorithm, which we call \emph{TCT}: first, use any off-the-shelf federated algorithm~\citep[such as FedAvg,][]{mcmahan2017communication} to train a deep model to extract useful features; then, compute a convex approximation of the deep model using its empirical Neural Tangent Kernel (eNTK)~\citep{jacot2018neural,lee2019wide,fortntk,long2021properties,wei2022more}, and use gradient correction methods such as SCAFFOLD to train the final model. Effectively, the second-stage features freeze the features learned in the first stage and fit a linear model over them. We show that this simple strategy is highly performant on a variety of tasks and models---we obtain accuracy gains up to $36$\% points on FMNIST with a CNN, $37$\% points on CIFAR10 with ResNet18-GN, and $16$\% points on CIFAR100 with ResNet18-GN. Further, its convergence remains unaffected even by extreme data heterogeneity. Finally, we show that given a pre-trained model, our method completely closes the gap between centralized and federated methods.

\section{Related Work}\label{sec:relatedwork}
\paragraph{Federated learning.} There are two main motivating scenarios for federated learning (FL). The first is where internet service companies (e.g., Google, Facebook, Apple, etc.) want to train machine learning models over their users' data, but do not want to transmit raw personalized data away from user devices~\citep{ramaswamy2020training, bonawitz2021federated}. 
This is the setting of \emph{cross-device} federated learning and is characterized by an extremely large number of unreliable clients, each of whom has very little data and the collections of data are assumed to be homogeneous~\citep{kairouz2019federated, bonawitz2019towards, karimireddy2020mime, bonawitz2021federated}. The second motivating scenario is when valuable data is split across different organizations, each of whom is either protected by privacy regulation or is simply unwilling to share their raw data. Such ``data islands'' are common among hospital networks, financial institutions, autonomous-vehicle companies, etc. This is known as \emph{cross-silo} federated learning and is characterized by a few highly reliable clients, who potentially have extremely diverse data. In this work, we focus on the latter scenario.

\vspace{-0.1in}
\paragraph{Metrics in FL.} FL research considers numerous metrics, such as fairness across users~\citep{mohri2019agnostic, li2019fair, shi2021survey}, formal security and privacy guarantees~\citep{bonawitz2017practical, ramaswamy2020training, fung2018mitigating, mothukuri2021survey}, robustness to corrupted agents and corrupted training data~\citep{blanchard2017machine,so2020byzantine,fang2020local,karimireddy2022byzantine,he2022byzantine}, preventing backdoors at test time~\citep{bagdasaryan2020backdoor,sun2019can,wang2020attack,lyu2020threats}, etc. While these concerns are important, the main goal of FL (and our work) is to achieve high accuracy with minimal communication~\citep{mcmahan2017communication}. Clients are typically geographically separated yet need to communicate large deep learning models over unoptimized ad-hoc networks~\citep{kairouz2019federated}. 
Finally, we focus on the setting where all users are interested in training the same model over the combined dataset. This is in contrast to model-agnostic protocols~\citep{lin2020ensemble, ozkara2021quped,afonin2021towards} or personalized federated learning~\citep{deng2020adaptive,fallah2020personalized,wu2020personalized,collins2021exploiting,kulkarni2020survey,chayti2022optimization}. Finally, we focus on minimizing the number of rounds required. Our approach can be combined with communication compression, which reduces bits sent per round~\citep{suresh2017distributed,alistarh2017qsgd,haddadpour2021federated,stich2020error}.

\vspace{-0.1in}
\paragraph{Federated optimization.} Algorithms for FL proceed in rounds. In each round, the server sends a model to the clients, who partially train this model using their local compute and data. The clients send these partially trained models back to the server who then aggregates them, finishing a round. FedAvg~\citep{mcmahan2017communication}, which is the de facto standard FL algorithm, uses SGD to perform local updates on the clients and aggregates the client models by simply averaging their parameters. Unfortunately, however, FedAvg has been observed to perform poorly when faced with data heterogeneity across the clients~\cite[etc.]{hsieh2020non,li2020federated,karimireddy2020scaffold,reddi2021adaptive,wang2020tackling,acar2021federated,li2021model,afonin2021towards,wang2021field,du2022flamby}. 
Theoretical investigations of this phenomenon~\citep{karimireddy2020scaffold,woodworth2020minibatch} showed that this was a result of \emph{gradient heterogeneity} across the clients. Consider FedAvg initialized with the globally optimal model. If this model is not also optimal for each of the clients as well, the local updates will push it away from the global optimum. Thus, convergence would require a careful tuning of hyper-parameters. To overcome this issue, SCAFFOLD~\citep{karimireddy2020scaffold} and FedDyn~\citep{acar2021federated} propose to use control variates to correct for the biases of the individual clients akin to variance reduction~\citep{johnson2013accelerating,defazio2014saga}. This \emph{gradient correction} is applied in every local update by the client and provably nullifies the effect of gradient heterogeneity~\citep{karimireddy2020scaffold,mishchenko2022proxskip,chayti2022optimization}. However, as we show here, such methods are insufficient to overcome high data heterogeneity especially for deep learning. Other, more heuristic approaches to combat {gradient heterogeneity} include using a regularizer~\citep{li2020federated} and sophisticated server aggregation strategies such as momentum~\citep{hsu2019measuring,wang2019slowmo,lin2021quasi} or adaptivity~\citep{reddi2021adaptive,karimireddy2020mime,charles2021large}. 

A second line of work pins the blame on performance loss due to averaging nonconvex models. To overcome this, \citet{singh2020model,yu2021fed2} propose to learn a mapping between the weights of the client models before averaging, \citet{afonin2021towards} advocates a functional perspective and replaces the averaging step with knowledge distillation, and \citet{wang2019adaptive,li2021model,tan2021fedproto} attempt to align the internal representations of the client models. However, averaging is unlikely to be the only culprit since FedAvg does succeed under low heterogeneity, and averaging nonconvex models can lead to improved performance~\citep{izmailov2018averaging,wortsman2022model}.

\paragraph{Neural Tangent Kernels (NTK) and neural network linearization.} NTK was first proposed to analyze the limiting behavior of infinitely wide networks~\citep{jacot2018neural,lee2019wide}. While NTK with MSE may be a bad approximation of real-world finite networks in general~\citep{goldblum2019truth}, it approximates the fine-tuning of a pre-trained network well~\citep{mu2020gradients}, especially with some minor modifications~\citep{achille2021lqf}. That is, NTK cannot capture feature learning but does capture how a model utilizes learnt features better than last/mid layer activations.

\begin{figure*}[t!]
\subcapcentertrue
  \begin{center}
    \subfigure[\label{fig:intro-fig-1}Using linear model.]{
        \includegraphics[width=0.46\textwidth]{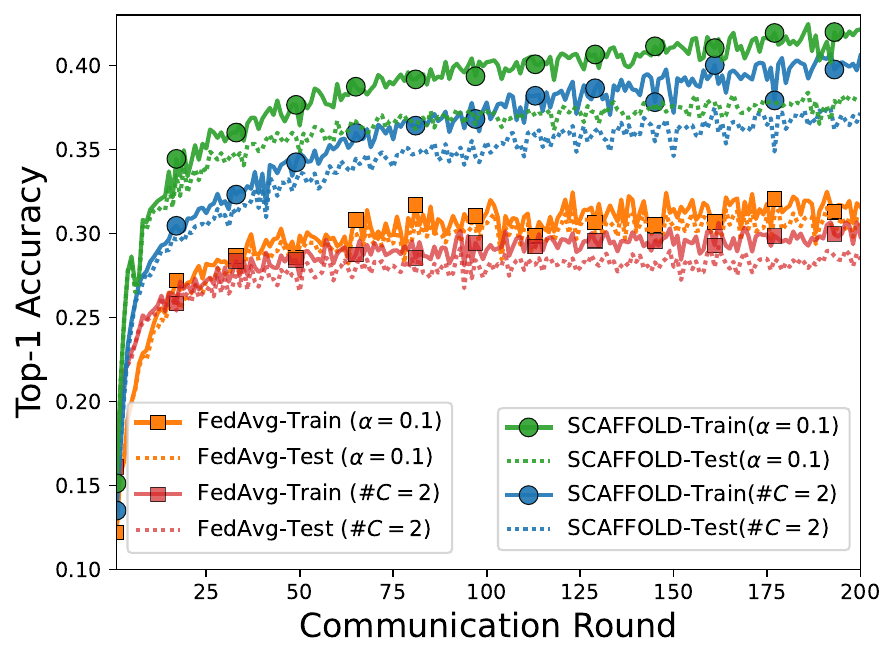}
    }
    \subfigure[\label{fig:intro-fig-2}Using ResNet-18.]{
        \includegraphics[width=0.46\textwidth]{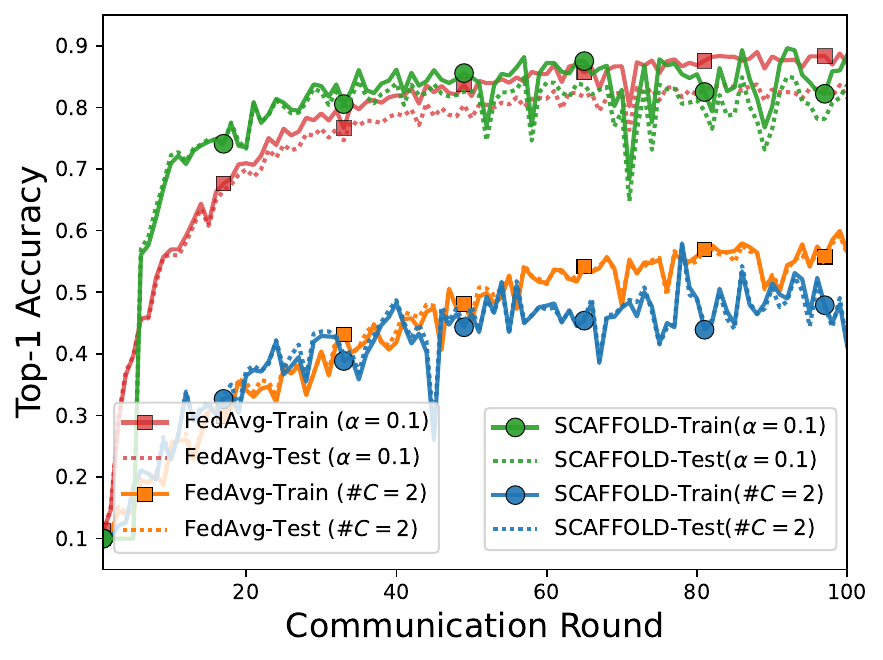}
    }
    \vskip -0.15in
    \caption{Performance of FedAvg and SCAFFOLD on CIFAR10 when data are split among ten clients in two ways (\texttt{\#C=2} and $\alpha$=0.1). The \texttt{\#C=2} split is more non-i.i.d. than the $\alpha$=0.1 split. For convex problems (left), gradient correction methods such as SCAFFOLD are relatively unaffected by data heterogeneity, and consistently outperform FedAvg. However, for nonconvex problems (right), FedAvg and SCAFFOLD perform very similarly and both are strongly negatively affected by data heterogeneity.
    }
    \label{fig:intro-fig}
  \end{center}
  \vskip -0.3in
\end{figure*}

\vspace{-0.05in}
\section{The Effect of Nonconvexity}\label{sec:model_specification}

In this section, we investigate the poor performance of FedAvg~\citep{mcmahan2017communication} and SCAFFOLD~\citep{karimireddy2020scaffold} empirically in the setting of deep neural networks, focusing on image classification with a ResNet-18. 
To construct our federated learning setup, we split the CIFAR-10 dataset in a highly heterogeneous manner among ten clients. We either assign each client two classes (denoted by \texttt{\#C=2}) or distribute samples according to a Dirichlet distribution with $\alpha=0.1$ (denoted by $\alpha$=0.1). For more details, see Section~\ref{subsec:setup}.

\vspace{-0.1in}
\paragraph{Insufficiency of gradient correction methods.}
Current theoretical work \citep[e.g.,][]{karimireddy2020scaffold,reddi2021adaptive,acar2021federated,wang2022unreasonable} attributes the slowdown from data heterogeneity to the individual clients having varying local optima. If no single model is simultaneously optimal for all clients, then the updates of different clients can compete with and distort each other, leading to slow convergence. This tension is captured by the variance of the updates across the clients~\citep[client gradient heterogeneity, see][]{wang2021field}. Gradient correction methods such as SCAFFOLD~\citep{karimireddy2020scaffold} and FedDyn~\citep{acar2021federated} explicitly correct for this and are provably unaffected by gradient heterogeneity for both convex and nonconvex losses.

These theoretical predictions are aligned with the results of Figure~\ref{fig:intro-fig-1}, where the loss landscape is convex: SCAFFOLD is relatively unaffected by the level of heterogeneity and consistently outperforms FedAvg. In particular, performance is largely dictated by the algorithm and not the data distributions. This shows that client gradient heterogeneity captures the difficulty of the problem well. On the other hand, when training a ResNet-18 model with nonconvex loss landscape, Figure~\ref{fig:intro-fig-2} shows that both FedAvg and SCAFFOLD suffer from data heterogeneity. This is despite the theory of gradient correction applying to both convex and nonconvex losses.  Further, the train and test accuracies in Figure~\ref{fig:intro-fig-2} match quite closely, suggesting that the failure lies in optimization (not fitting the training data) rather than generalization. Thus, while the current theory makes no qualitative distinctions between convex and nonconvex convergence, the practical behavior of algorithms in these settings is very different. Such differences between theoretical predictions and practical reality suggests that black-box notions such as gradient heterogeneity are insufficient for capturing the difficulty of training deep models.

\begin{table}[!t]
    \centering
    \caption{Feature learning by FedAvg. We report the test accuracy of a ResNet-18 after (centralized) retraining of the last $\ell$ layers on CIFAR10. The earlier $(7-\ell)$ layers are frozen to either random initialization or to the weights of a FedAvg-trained model. The difference measures utility of the $(7-\ell)$ layers learnt by FedAvg. The baseline FedAvg model without additional training gets $56.9$\% accuracy. 
    We see that all layers of the FedAvg model contain useful information.
    }
    \vspace{0.1in}
    \begin{tabular}{@{}l l l l@{}r@{}}
    \toprule
    Layers retrained & \multicolumn{1}{c}{Accuracy (\%)} & \multicolumn{1}{c}{Accuracy (\%)} & \multicolumn{2}{c}{Improvement (\%)}\\
                    & \multicolumn{1}{c}{Random init}  & \multicolumn{1}{c}{FedAvg init} & \multicolumn{2}{c}{(FedAvg - Random)}\\
    \midrule
    1/7 last layer & 35.37 & 77.93 & \barplot{35.37}{77.93}\\
    2/7 last layers & 67.33 & 87.04& \barplot{67.33}{87.04}\\
    3/7 last layers & 80.18 & 89.28 & \barplot{80.18}{89.28}\\
    4/7 last layers & 88.03 & 90.57& \barplot{88.03}{90.57}\\
    5/7 last layers & 91.34 & 91.61& \barplot{91.34}{91.61}\\
    6/7 last layers & 91.78 & 91.91& \barplot{91.78}{91.91}\\
    \bottomrule
    \end{tabular}
    \label{tab:model_specification_table1}
    \vspace{-0.15in}
\end{table}
    

\vspace{-0.1in}
\paragraph{Ease of feature learning.}
We now dive into how a ResNet-18 trained with FedAvg ($56.9$\% accuracy) differs from the centralized baseline ($91.9$\% accuracy). We first apply linear probing to the FedAvg model (i.e., retraining with all but the output layer frozen). Note that this is equivalent to (convex) logistic regression over the last-layer activations. This simple procedure produces a striking jump from $56.9$\% to $77.9$\% accuracy. Thus, of the $35$\% gap in accuracy between the FedAvg and centralized models, $21$\% may be attributed to a failure to optimize the linear output layer. We next extend this experiment towards probing the information content of other layers.

Given a FedAvg-trained model, we can use centralized training to retrain only the last $\ell$ layers while keeping the rest of the $(7-\ell)$ layers (or ResNet blocks) frozen. We can also perform this procedure starting from a randomly initialized model. The performance difference between these two models can be attributed to the information content of the frozen $(7-\ell)$ layers of the FedAvg model. Table~\ref{tab:model_specification_table1} summarizes the results of this experiment. The large difference in accuracy (up to $42.6$\%) indicates the initial layers of the FedAvg model have learned useful features. There continues to be a gap between the FedAvg features and random features in the earlier layers as well,\footnote{The significant decrease in the gap as we go down the layers may be because of the skip connections in the lower ResNet blocks which allow the random frozen layers to be sidestepped. This underestimates the true utility and information content in the earlier FedAvg layers.} meaning that all layers of the FedAvg model learn useful features. We conjecture this is because from the perspective of earlier layers which perform simple edge detection, the tasks are independent of labels and the clients are i.i.d. However, the higher layers are more specialized and the effect of the heterogeneity is stronger.

\section{Method}\label{sec:method}
Based on the observations in Section~\ref{sec:model_specification}, we propose \textit{train-convexify-train} (TCT) as a method for overcoming data heterogeneity when training deep models in a federated setting.
Our high-level intuition is that we want to leverage both the features learned from applying FedAvg to neural networks and the effectiveness of \emph{convex} federated optimization. We thus perform several rounds of ``bootstrap'' FedAvg to learn features before solving a convexified version of the original optimization problem.

\subsection{Computing the Empirical Neural Tangent Kernel}\label{subsec:method-eNTK}
To sidestep the challenges presented by nonconvexity, we describe how we approximate a neural network by its ``linearization.'' Given a neural network $f(\,\cdot\,;\theta_0)$ with weights $\theta_0\in\mathbb{R}^P$ mapping inputs $x\in\mathbb{R}^D$ to $\mathbb{R}^C$, we replace it by its \emph{empirical neural tangent kernel (eNTK)} approximation at $\theta_0$ given by 
$$f(x; \theta)\approx f(x; \theta_0) + (\theta - \theta_0)^\top\frac{\partial}{\partial \theta} f(x; \theta_0),$$ at each $x\in\mathbb{R}^D$. 
Under this approximation, $f(x; \theta)$ is a linear function of the ``feature vector'' \smash{$(f(x; \theta_0), \frac{\partial}{\partial \theta} f(x; \theta_0))$} and the original nonconvex optimization problem becomes (convex) linear regression with respect to these features.\footnote{For classification problems, we one-hot encoded labels and fit a linear model using squared loss.}

To reduce the computational burden of working with the eNTK approximation, we make two further approximations: First, we randomly reinitialize the last layer of $\theta_0$ and only consider \smash{$\frac{\partial}{\partial \theta} f(x; \theta_0)$} with respect to a single output logit. Over the randomness of this reinitialization, $\mathbb{E}[f(x; \theta_0)] = 0$. Moreover, given the random reinitialization, all the output logits of $f(x; \theta_0)$ are symmetric. These observations mean each data point $x$ can be represented by a $P$-dimensional feature vector \smash{$\frac{\partial}{\partial \theta} f_1(x; \theta_0)$}, where $f_1(\,\cdot\,;\theta_0)$ refers to the first output logit. 
Then, we apply a dimensionality reduction by subsampling $p$ random coordinates from this $P$-dimensional featurization.\footnote{That such representations empirically have low effective dimension due to fast eigenvalue decay~\citep[see, e.g.,][]{wei2022more} means that such a random projection approximately preserves the geometry of the data points~\cite{avron17sharper, zancato2020predicting}. For all of our experiments, we set $p=100,000$.}  In our setting, this sub-sampling has the added benefit of reducing the number of bits communicated per round.

In summary, we transform our original (nonconvex) optimization problem over a neural network initialized at $\theta_0$ into a convex optimization problem in three steps: (i) reinitialize the last layer of $\theta_0$; (ii) for each data point $x$, compute the gradient \smash{$\phi_{\mathrm{eNTK}}(x; \theta_0)\coloneqq\frac{\partial}{\partial \theta} f_1(x; \theta_0)$}; (iii) subsample the coordinates of $\phi_{\mathrm{eNTK}}(x; \theta_0)$ for each $x$ to obtain a reduced-dimensionality eNTK representation. Let $\mathscr{S}\colon\mathbb{R}^{P} \rightarrow \mathbb{R}^{p}$ denote this subsampling operation. Finally, we solve the resulting linear regression problem over these eNTK representations.\footnote{Given a fitted linear model with weights \smash{${W}\in\mathbb{R}^{p\times C}$}, the prediction at $x$ is \smash{$\arg\max_{j}[W^{\top}\mathscr{S}(\phi_{\mathrm{eNTK}}(x))]_j$}.}

\subsection{Convexifying Federated Learning via eNTK Representations}
The eNTK approximation lets us convexify the neural net optimization problem: following Section~\ref{subsec:method-eNTK}, we may extract (from a model trained with FedAvg) eNTK representations of inputs from each client. It remains to fit an overparameterized linear model using these eNTK features in a federated manner.  For ease of presentation, we denote the subsampled eNTK representation of input $x$ by $z\in\mathbb{R}^{p}$, where $p$ is the eNTK feature dimension after subsampling. 
We use $z_{i}^{k}$ to represent the eNTK feature of the $i$-th sample from the $k$-th client. Then, for $K$ the number of clients, $Y_i^{k}$ the one-hot encoded labels, $n_k$ the number of data points of the $k$-th client, $n:=\smash{\sum_{k\in[K]}n_k}$ the number of data points across all clients, and $p_k:=n_k/n$,
we can approximate the nonconvex neural net optimization problem by the convex linear regression problem
\begin{equation}\label{eq:eNTK-linear-regression}
    \min_{W} L(W) := \sum_{k=1}^{K}p_k\cdot L_k(W),\qquad\text{where}\quad L_k(W) := {\frac{1}{n_k}\sum_{i=1}^{n_k}\|W^{\top}z_i^{k} - Y_{i}^{k}\|_{2}^{2}}.
\end{equation}
To obtain the eNTK representation $z$ of an input $x$, we take $\theta_0$ in Section~\ref{subsec:method-eNTK} to be the weights of a model trained with FedAvg. As we will show in Section~\ref{sec:exp}, the convex reformulation in Eq.~\eqref{eq:eNTK-linear-regression} significantly reduces the number of communication rounds needed to find an optimal solution.

\subsection{{\textbf{T}rain-\textbf{C}onvexify-\textbf{T}rain} (TCT)}
We now present our algorithm {train-convexify-train} (TCT), with convexification done via the neural tangent kernel, for federated optimization. 

\vspace{0.05in}
\begin{tcolorbox}
\begin{center}
    {\large \textbf{TCT} ---  train-convexify-train with eNTK representations}
\end{center}
\begin{itemize}
    \item \textbf{Stage 1:} \textit{Extract eNTK features from a FedAvg-trained model.} FedAvg is first used to train the model for $T_1$ communication rounds. Let $\theta_{T_1}$ denote the model weights after these $T_1$ rounds. Then, each client locally computes subsampled eNTK features, i.e., $z_i^{k}=\mathscr{S}(\phi_{\mathrm{eNTK}}(x_i^k; \theta_{T_1}))$ for $k\in[K]$ and $i\in[n_k]$.
    \item \textbf{Stage 2:} \textit{Decentralized linear regression with gradient correction.} Given samples $\{(z_i^{k}, Y_i^{k})\}_{i=1}^{n_k}$ on each client $k$, first normalize the eNTK inputs of all clients with a single communication round.\footnote{For every feature in the eNTK representation, subtract the mean and scale to unit variance.} Then, solve the linear regression problem defined in Eq.~\eqref{eq:eNTK-linear-regression} by SCAFFOLD with local learning rate $\eta$ and local steps $M$.\footnote{The detailed description of SCAFFOLD for solving linear regression problems can be found in Algorithm~\ref{Algorithm:Scaffold-eNTK}, Appendix~\ref{sec:appendix-algorithm-details}. It has the same communication and computation cost as FedAvg.}
\end{itemize}
\end{tcolorbox} 

\noindent
To motivate TCT, recall that in Section~\ref{sec:model_specification} we found that FedAvg learns ``useful'' features despite its poor performance, especially in the earlier layers. By taking an eNTK approximation, TCT optimizes a convex approximation while using information from \emph{all} layers of the model. 
Empirically, we find that these extracted eNTK features significantly reduce the number of communication rounds needed to learn a performant model, even with data heterogeneity.

\section{Experiments}\label{sec:exp}
We now study the performance of TCT for the decentralized training of deep neural networks in the presence of data heterogeneity. We compare TCT to state-of-the-art federated learning algorithms on three benchmark tasks in federated learning. For each task, we apply these algorithms on client data distributions with varying degrees of data heterogeneity. We find that our proposed approach significantly outperforms existing algorithms when clients have highly heterogeneous data across all tasks. For additional experimental results and implementation details,  see Appendix~\ref{sec:appendix-additional-exp}. Our code is available at \url{https://github.com/yaodongyu/TCT}.

\subsection{Experimental Setup}\label{subsec:setup}

\noindent
\textbf{Datasets and degrees of data heterogeneity.}
We assess the performance of federated learning algorithms on the image classification tasks FMNIST~\citep{xiao2017fashion}, CIFAR10, and CIFAR100~\citep{krizhevsky2009learning}. FMNIST and CIFAR10 each consist of 10 classes, while CIFAR100 includes images from 100 classes. There are 60,000 training images in FMNIST, and 50,000 training images in CIFAR10/100.

To vary the degree of data heterogeneity, we follow the setup of \citet{li2021federated}. We consider two types of non-i.i.d.\ data distribution: \emph{(i) Data heterogeneity sampled from a symmetric Dirichlet distribution with parameter $\alpha$}~\citep{lin2020ensemble, wang2020tackling}. 
That is, we sample $p_{c}\sim\mathrm{Dir}_{K}(\alpha)$ from a $K$-dimensional symmetric Dirichlet distribution and assign a $p_{c}^{k}$-fraction of the class $c$ samples to client $k$. (Smaller $\alpha$ corresponds to more heterogeneity.) \emph{(ii) Clients get samples from a fixed subset of classes}~\citep{mcmahan2017communication}. 
That is, each client is allocated a subset of classes; then, the samples of each class are split into non-overlapping subsets and assigned to clients that were allocated this class. We use $\texttt{\#C}$ to denote the number of classes allocated to each client. For example, $\texttt{\#C=2}$ means each client has samples from 2 classes. To allow for consistent comparisons, all of our experiments are run with $10$ clients.

\vspace{-0.1in}
\paragraph{Models.} For FMNIST, we use a convolutional neural network with ReLU activations consisting of two convolutional layers with max pooling followed by two fully connected layers (SimpleCNN). For CIFAR10 and CIFAR100, we mainly consider an 18-layer residual network~\citep{he2016deep} with 4 basic residual blocks (ResNet-18). In Appendix~\ref{subsec:appendix-other-arch}, we present experimental results for other architectures. 

\vspace{-0.1in}
\paragraph{Algorithms and training schemes.} We compare TCT to state-of-the-art federated learning algorithms, focusing on the widely-used algorithms {FedAvg}~\citep{mcmahan2017communication}, {FedProx}~\citep{li2020federated}, and {SCAFFOLD}~\citep{karimireddy2020scaffold}. (For comparisons to additional algorithms, see Appendix~\ref{subsec:appendix-more-baselines}.) Each client uses SGD with {weight decay $10^{-5}$} and batch size $64$ by default. 
For each baseline method, we run it for $200$ total communication rounds using $5$ local training epochs with local learning rate selected from $\{0.1, 0.01, 0.001\}$ by grid search. For TCT, we run $100$ rounds of FedAvg in Stage 1 following the above and use $100$ communication rounds in Stage 2 with $M=500$ local steps and local learning rate $\eta=5\cdot 10^{-5}$.

\begin{table*}[t]
	\centering
	\caption{{The top-1 accuracy ($\%$) of our algorithm (TCT) vs.\ state-of-the-art federated learning algorithms evaluated on FMNIST, CIFAR10, and CIFAR100. We vary the degree of data heterogeneity by controlling the $\alpha$ parameter of the symmetric Dirichlet distribution $\mathrm{Dir}_K(\alpha)$ and the $\texttt{\#C}$ parameter for assigning how many labels each client owns. Higher accuracy is better. The highest top-1 accuracy in each setting is highlighted in \textbf{bold}. } 
	}
	\vspace{0.75em}
	\label{tab:main_table}
	    \footnotesize
		\begin{tabular}{ccccccccc}
			\toprule
			\multirow{1}{*}{\bf Datasets} & \multirow{1}{*}{\bf Architectures} &   \multirow{1}{*}{\bf Methods}                 & \multicolumn{4}{c}{\bf Non-i.i.d. degree}   
			  \\
			  \midrule
		\midrule
		\multirow{7}{*}{FMNIST}   & \multirow{7}{*}{SimpleCNN} & & $\texttt{\#C}=1$ & $\texttt{\#C}=2$  & $\alpha=0.1$ & 	$\alpha=0.5$   \\ 
			\cmidrule(lr){1-7} 
			  &   & {FedAvg}  & 35.10\%   & 85.18\%  & 86.18\% & 90.09\%    \\ 
			\tablewhitespace
			  &   & {FedProx}  & 50.04\%  & 84.91\%  & 86.31\%   & 89.77\%        \\
			  \tablewhitespace
			 
			  &   & {SCAFFOLD}  & 12.80\%   & 42.80\%   & 83.87\%   &  89.40\%     \\
			  \tablewhitespace
                &   & \textit{TCT}  & \textbf{86.32\%}   & \textbf{90.33\%}  & \textbf{90.78\%}  & \textbf{91.13\%} \\
                \cmidrule(lr){3-7} 
                &   & {\color{tab10_blue}\textit{Centralized}}  & \multicolumn{4}{c}{{\color{tab10_blue}91.40\%}} \\
            \midrule
		\midrule
		\multirow{7}{*}{CIFAR-10}   & \multirow{7}{*}{ResNet-18} &     & $\texttt{\#C}=1$ & $\texttt{\#C}=2$  & $\alpha=0.1$ & 	$\alpha=0.5$
			\\ 
			\cmidrule(lr){1-7} 
			  &   & {FedAvg} & 11.27\%  & 56.86\%  & 82.60\%      & 90.43\%          \\
			  \tablewhitespace 
			  &   & {FedProx} &      12.30\%  & 56.87\%  & 83.31\%  & 90.68\%        \\
			  \tablewhitespace
			  &   & {SCAFFOLD} &    10.00\% & 46.75\% & 80.46\%    & 90.72\%      \\
			  \tablewhitespace
                &   & \textit{TCT}   &  \textbf{49.92\%}  & \textbf{83.02\%} & \textbf{89.21\%}  & \textbf{91.10\%}  \\
                \cmidrule(lr){3-7} 
                &   & {\color{tab10_blue}\textit{Centralized}}  & \multicolumn{4}{c}{{\color{tab10_blue}91.90\%}} \\
            \midrule
            \midrule
			\multirow{7}{*}{CIFAR-100}   & \multirow{7}{*}{ResNet-18} &    &
			$\alpha=0.001$ & $\alpha=0.01$ & $\alpha=0.1$ & $\alpha=0.5$ 
			\\ 
			\cmidrule(lr){1-7} 
			  &   & {FedAvg} & 53.89\% & 54.22\%  & 63.49\%  & 67.65\%    \\ 
			\tablewhitespace
			  &   & {FedProx}  &  52.87\%  & 54.32\%  & 63.47\%  &  67.54\%    \\
			  \tablewhitespace
			  &   & {SCAFFOLD} &  49.86\%  & 54.07\%   & 65.67\%  &  \textbf{71.07\%}    \\
			  \tablewhitespace
                &   & \textit{TCT}  & \textbf{68.42\%}  & \textbf{69.07\%}  & \textbf{69.66\%}  & {69.68\%}   \\
                \cmidrule(lr){3-7} 
                &   & {\color{tab10_blue}\textit{Centralized}}  & \multicolumn{4}{c}{{\color{tab10_blue}73.61\%}} \\
			\bottomrule
		\end{tabular}
	\vspace{-.75em}
\end{table*}

\subsection{Main Results}

\begin{figure*}[t]
\subcapcentertrue
  \begin{center}
    \subfigure[\label{fig:main-fig-1} {FedAvg}.]{\includegraphics[width=0.325\textwidth]{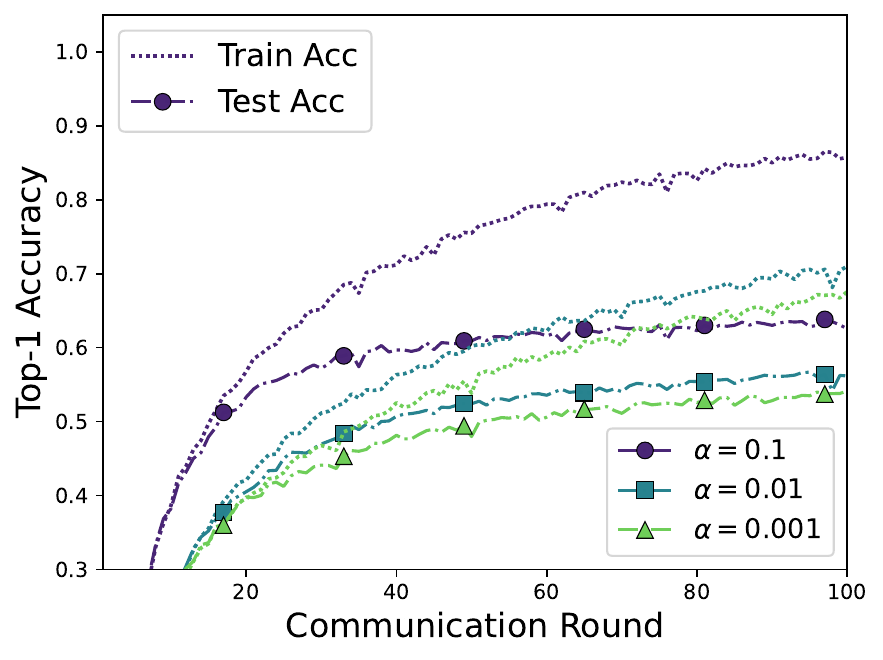}}
    \subfigure[\label{fig:main-fig-2}{SCAFFOLD}.]{\includegraphics[width=0.325\textwidth]{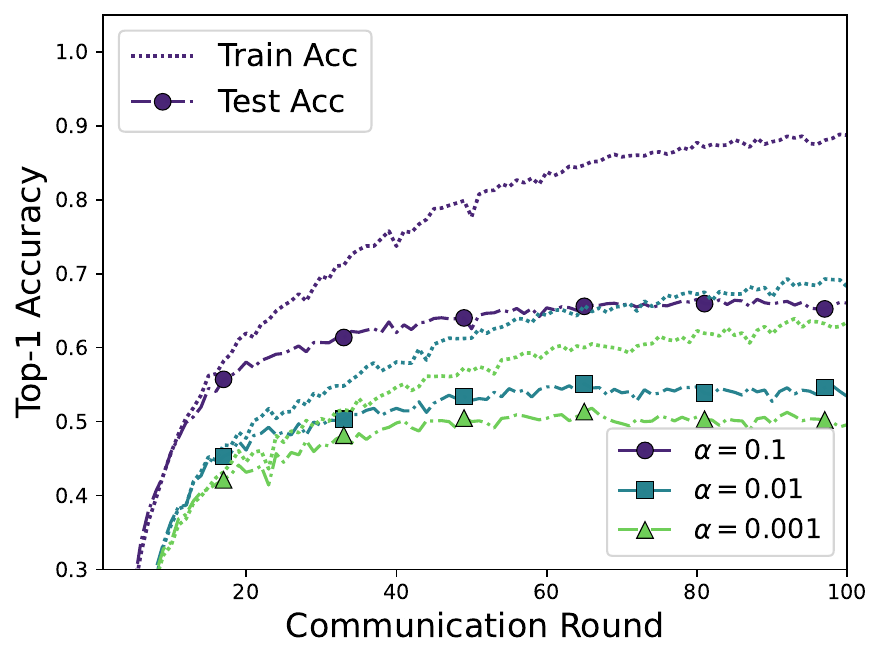}}
    \subfigure[\label{fig:main-fig-3} TCT.]{\includegraphics[width=0.325\textwidth]{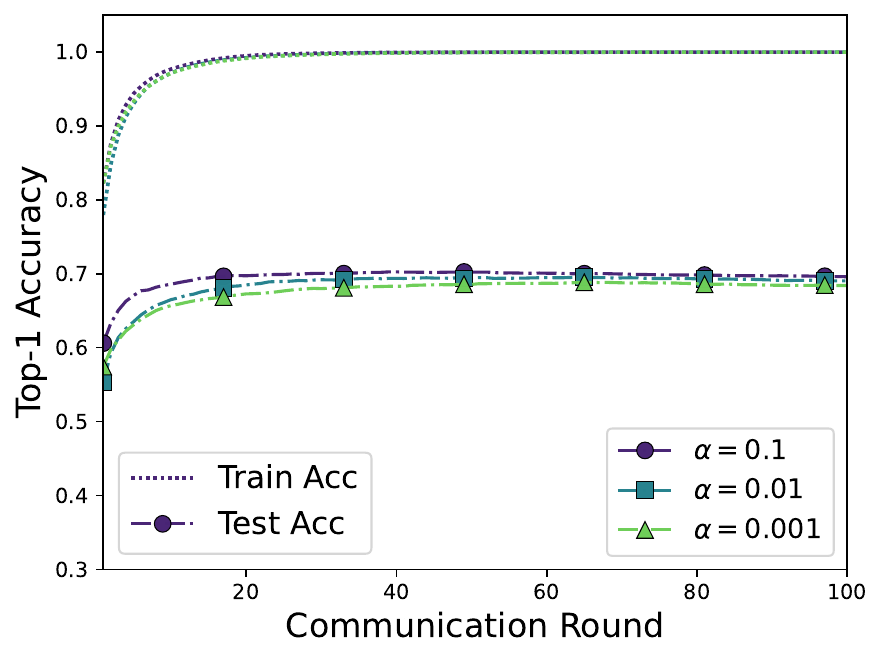}}
    \vskip -0.1in
    \caption{Training/test accuracy vs.\ communication round for {FedAvg} (left), {SCAFFOLD} (middle), and our algorithm TCT (right) on the CIFAR100 dataset with various degrees of non-iid-ness ($\mathrm{Dir}_{K}(\alpha)$ with $\alpha\in\{0.1, 0.01, 0.001\}$). Dotted lines represent the training accuracy, and dashdot lines with markers represent the test accuracy.  
    }
    \label{fig:main-fig}
  \end{center}
  \vskip -0.3in
\end{figure*}

Table~\ref{tab:main_table} displays the top-1 accuracy of all algorithm on the three tasks with varying degrees of data heterogeneity. We evaluated each algorithms on each task under four degrees of data heterogeneity. Smaller $\texttt{\#C}$ and $\alpha$ in Table~\ref{tab:main_table} correspond to higher heterogeneity.

We find that the existing federated algorithms all suffer when data heterogeneity is high across all three tasks. For example, the top-1 accuracy of {FedAvg} on CIFAR-10 is $56.86\%$ when $\texttt{\#C=2}$, which is much worse than the $90.43\%$ achieved in a more homogeneous setting (e.g. $\alpha=0.5$). 
In contrast, TCT achieves consistently strong performance, even in the face of high data heterogeneity.  More specifically, TCT achieves the best top-1 accuracy performance across all settings except CIFAR-100 with $\alpha=0.5$, where TCT does only slightly worse than {SCAFFOLD}. 

In absolute terms, we find that TCT is not affected much by data heterogeneity, with performance dropping by less than $1.5\%$ on CIFAR100 as $\alpha$ goes from $0.5$ to $0.001$. Moreover, our algorithm improves over existing methods by at least $15\%$ in the challenging cases, including FMNIST with $\texttt{\#C=1}$, CIFAR-10 with $\texttt{\#C=1}$ and $\texttt{\#C=2}$, and CIFAR-100 with $\alpha=0.01$ and $\alpha=0.001$. And, perhaps surprisingly, our algorithm still performs relatively well in the extreme non-i.i.d.\ setting where each client sees only a single class.

Figure~\ref{fig:main-fig} compares the performances of {FedAvg}, {SCAFFOLD}, and TCT in more detail on CIFAR100 dataset with different degrees of data heterogeneity. 
We consider the Dirichlet distribution with parameter $\alpha \in \{0.1, 0.01, 0.001\}$ and compare the training and test accuracy of these three algorithms. As shown in Figures~\ref{fig:main-fig-1} and \ref{fig:main-fig-2}, both {FedAvg} and {SCAFFOLD} struggle when data heterogeneity is high: for both algorithms, test accuracy drops significantly when $\alpha$ decreases. 
In contrast, we see from Figure~\ref{fig:main-fig-3} that TCT  maintains almost the same test accuracy for different $\alpha$. Furthermore, the same set of default parameters for our algorithm, including local learning rate and the number of local steps, is relatively robust to different levels of data heterogeneity.\vspace{-0.5em}

\subsection{Communication Efficiency} 
\begin{figure*}[t]
\subcapcentertrue
  \begin{center}
    \subfigure[\label{fig:commute-efficiency-1}FMNIST.]{\includegraphics[width=0.325\textwidth]{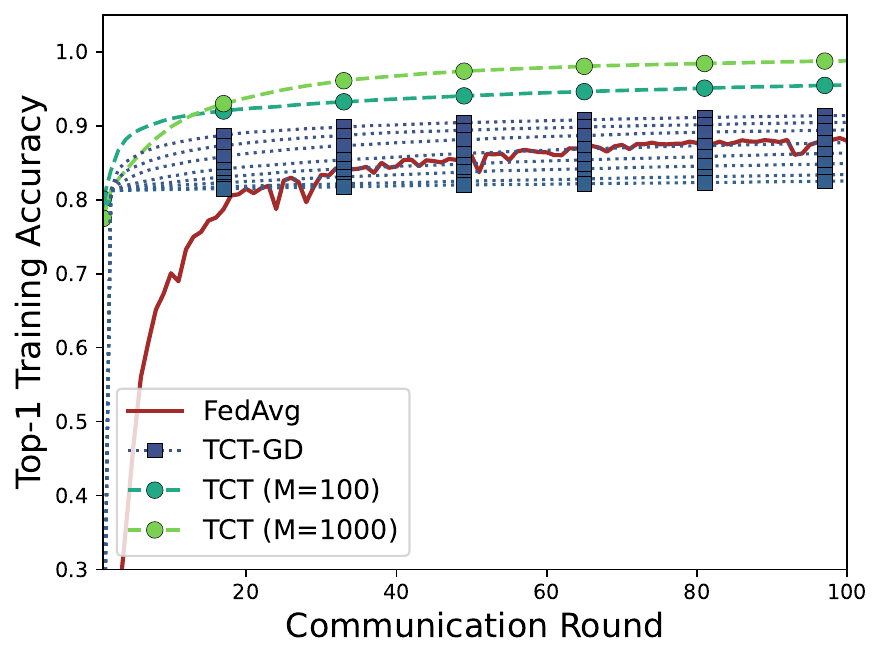}}
    \subfigure[\label{fig:commute-efficiency-2}CIFAR10.]{\includegraphics[width=0.325\textwidth]{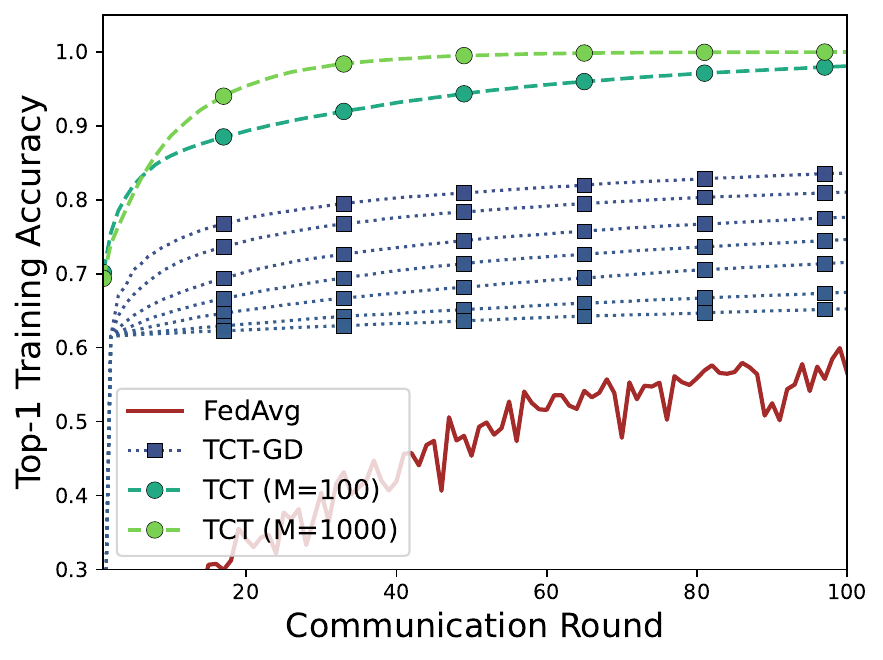}}
    \subfigure[\label{fig:commute-efficiency-3}CIFAR100.]{\includegraphics[width=0.325\textwidth]{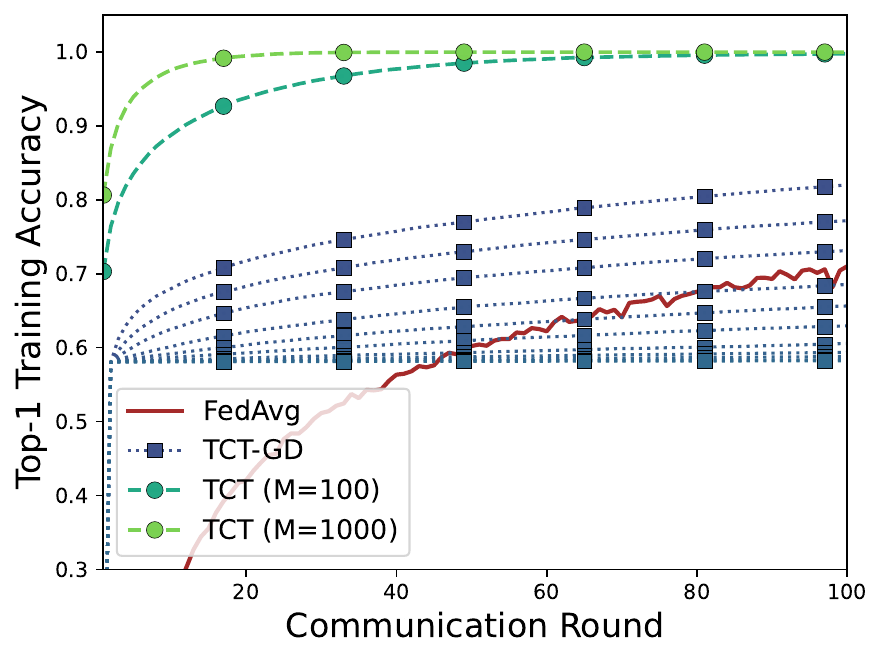}}
    \vskip -0.1in
    \caption{Training accuracy vs.\ communication round for full batch gradient descent (GD) and TCT on FMNIST-[\texttt{\#C=2}]~\textbf{(a)}, CIFAR10-[\texttt{\#C=2}]~\textbf{(b)}, and CIFAR100-[$\alpha=0.01$]~\textbf{(c)}. Each dotted line with square markers represents the training accuracy of GD with some learning rate. Dashed lines with circle markers represent the training accuracy of TCT with different numbers of local steps. We also include the training accuracy results of {FedAvg} with learning rate $\eta=0.1$. We use TCT-GD to denote the variant of TCT which replaces SCAFFOLD with GD in Stage 2.
    }
    \label{fig:commute-efficiency}
  \end{center}
  \vskip -0.3in
\end{figure*}

To understand the effectiveness of the local steps in our algorithm, we compare SCAFFOLD (used in TCT-Stage 2) to full batch gradient descent (GD) applied to the overparameterized linear regression problem in Stage 2 of TCT on these datasets. 
For our algorithm, we set local steps $M\in\{10^{2}, 10^{3}\}$ and use the default local learning rate. For full batch GD, we vary the learning rate from $10^{-5}$ to $10^{-1}$ and visualize the ones that do not diverge. 
The results are summarized in Figure~\ref{fig:commute-efficiency}. Each dotted line with square markers in Figure~\ref{fig:commute-efficiency} corresponds to full batch GD with some learning rate. 
Across all three datasets, our proposed algorithm consistently outperforms full batch GD.  Meanwhile, we find that more local steps for our algorithms lead to faster convergence across all settings. 
In particular, our algorithm converges within 20 communication rounds on CIFAR100 (as shown in Figure~\ref{fig:commute-efficiency-3}). These results suggest that our proposed algorithm can largely leverage the local computation and improve communication efficiency.

\begin{figure*}[t]
\subcapcentertrue
  \begin{center}
    \subfigure[\label{fig:ablation-fedavg-scaffold-1}FMNIST.]{\includegraphics[width=0.325\textwidth]{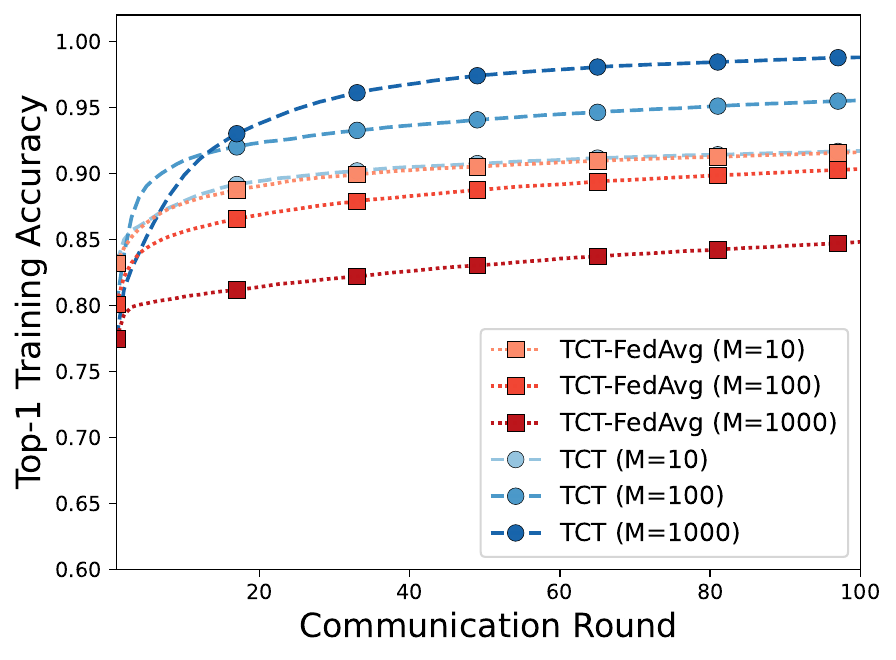}}
    \subfigure[\label{fig:ablation-fedavg-scaffold-2}CIFAR10.]{\includegraphics[width=0.325\textwidth]{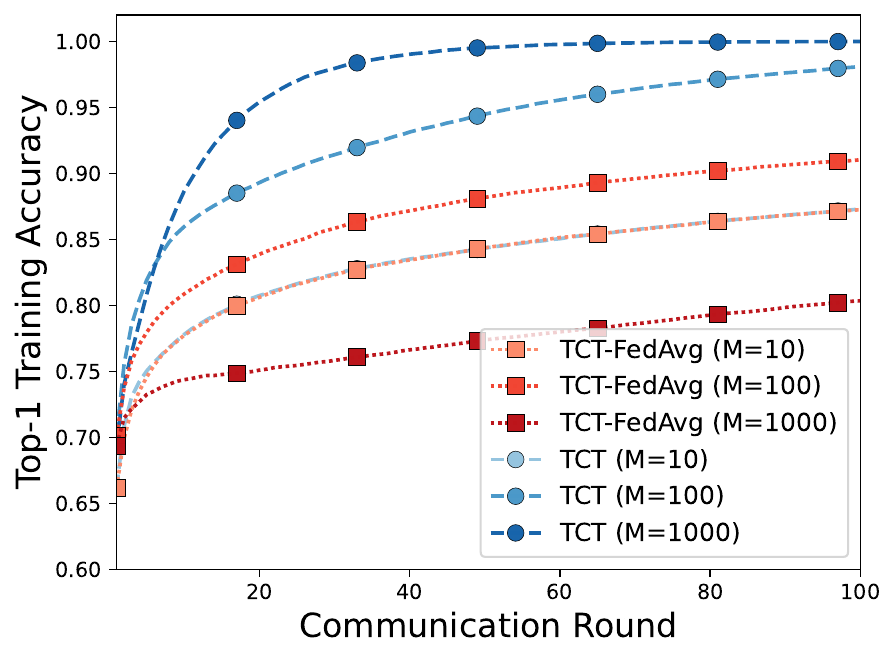}}
    \subfigure[\label{fig:ablation-fedavg-scaffold-3}CIFAR100.]{\includegraphics[width=0.325\textwidth]{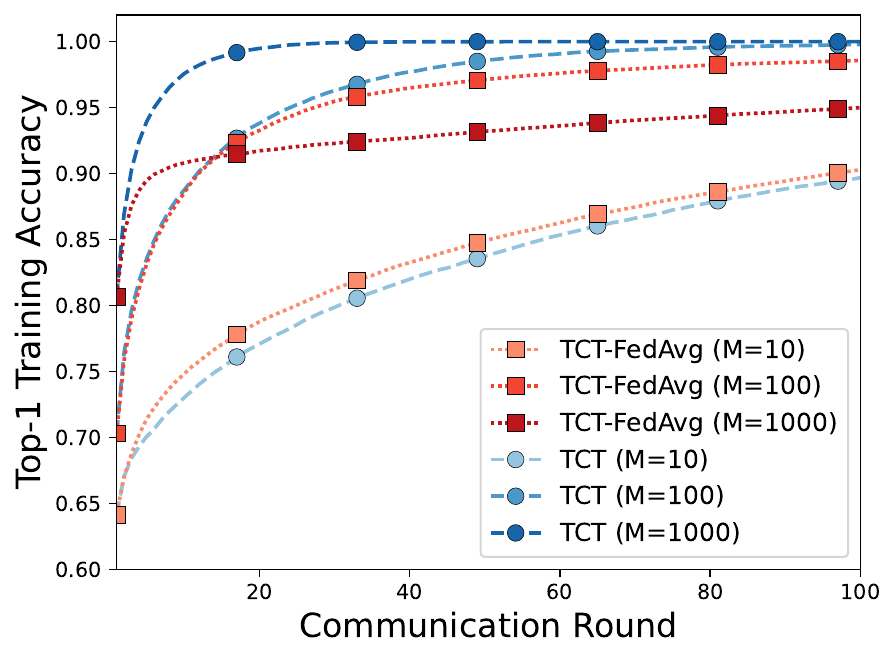}}
    \vskip -0.05in
    \caption{Comparing TCT to TCT-FedAvg for solving the overparameterized linear regression problem on \textbf{(a)}~FMNIST-[\texttt{\#C=2}], \textbf{(b)}~CIFAR10-[\texttt{\#C=2}], and \textbf{(c)}~CIFAR100-[$\alpha=0.01$]. We use TCT-FedAvg to denote a variant of TCT that uses FedAvg instead of SCAFFOLD to perform linear regression in TCT-Stage 2.   Dotted red lines with square markers represent the training accuracy of TCT-FedAvg with different numbers of local steps. Dashed blue red lines with circle markers represent the training accuracy of TCT with different numbers of local steps. A darker color means more local steps.
    }
    \label{fig:ablation-fedavg-scaffold}
  \end{center}
  \vskip -0.2in
\end{figure*}

\subsection{Ablations} 
\textbf{Gradient correction.} We investigate the role of gradient correction when solving overparameterized linear regression  with eNTK features in TCT.  We compare SCAFFOLD (used in TCT) to {FedAvg} on solving the regression problems and summarize the results in Figure~\ref{fig:ablation-fedavg-scaffold}. We use the default local learning rate and consider three different numbers of local steps for both algorithms, i.e., $M\in\{10, 100, 1000\}$. 
As shown in Figure~\ref{fig:ablation-fedavg-scaffold}, our approach largely outperforms {FedAvg} when the number of local steps is large ($M\geq100$) across three datasets. We also find that the performance of {FedAvg} can even degrade when the number of local steps increases. For example, {FedAvg} with $M=1000$ performs the worst across all three datasets. In contrast to {FedAvg}, SCAFFOLD converges faster when the number of local steps increases. These observations highlight the importance of gradient correction in our algorithm.

\begin{figure*}[t]
\subcapcentertrue
  \begin{center}
    \subfigure[\label{fig:method-fig-1}Effect of {FedAvg} communication rounds in Stage 1.]{\includegraphics[width=0.56\textwidth]{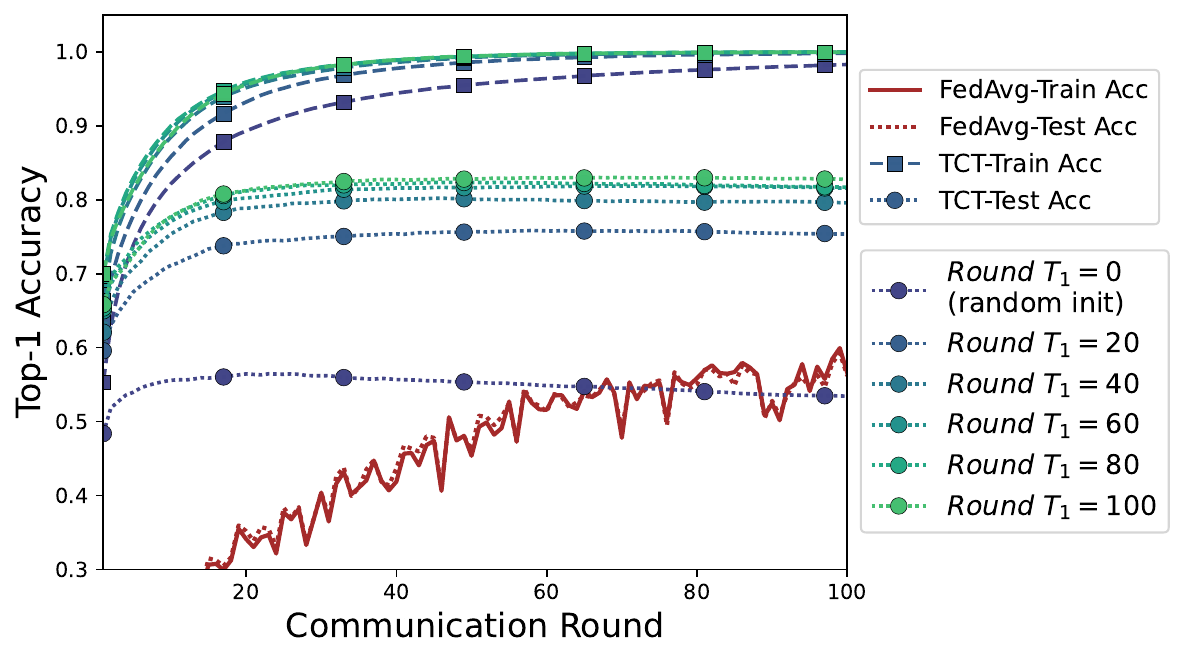}}
    \subfigure[\label{fig:method-fig-2}Effect of normalization.]{\includegraphics[width=0.42\textwidth]{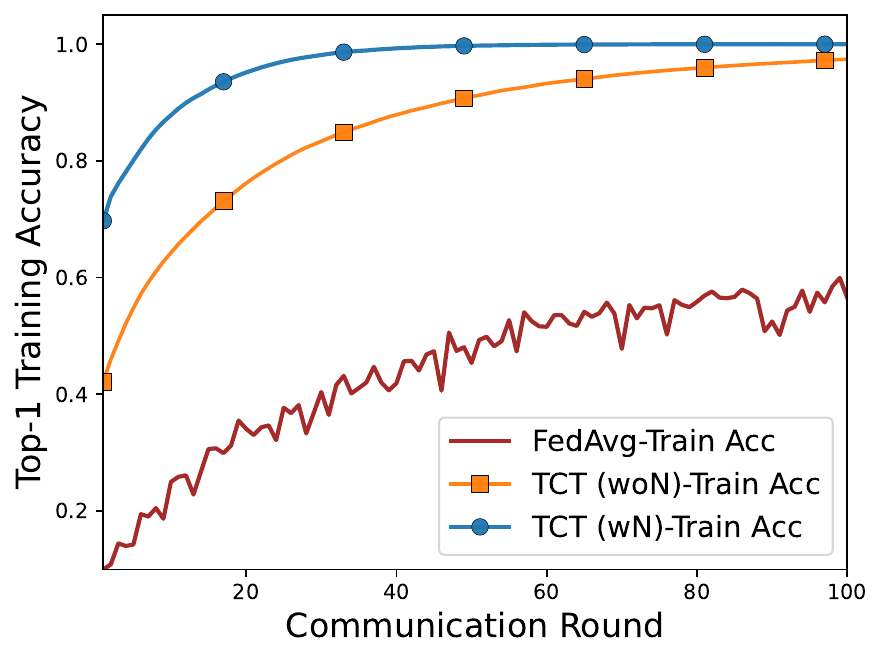}}
    \vskip -0.05in
    \caption{ \textbf{(a).} We evaluate TCT on using checkpoints save at different communication rounds $T_1$ in Stage 1. $T_1=0$ corresponds to the randmon initialized model weights scenario (without {FedAvg} training). Dash lines with square markers represent the training accuracy, and dotted lines with circle makers represent the test accuracy. \textbf{(b).} We study the effect of pre-conditioning on TCT. TCT~(wN) corresponds to the setting where eNTK features are normalized, and TCT~(woN) corresponds to the without normalization step setting. 
    }
    \label{fig:method-fig}
  \end{center}
  \vskip -0.2in
\end{figure*}

\vspace{-0.1in}
\paragraph{Model weights for computing eNTK features.} To understand the impact of the model weights trained in {Stage 1} of TCT, we evaluate TCT run with different $T_1$ parameters. We consider $T_1 \in \{0, 20, 40, 60, 80, 100\}$, where $T_1=0$ corresponds to randomly initialized weights. From Figure~\ref{fig:method-fig-1}, we find that weights after FedAvg training are much more effective than weights at random initialization. 
Specifically, without FedAvg training, the eNTK (at random initialization) performs worse than standard FedAvg. In contrast, TCT significantly outperforms FedAvg by a large margin (roughly $20\%$ in test accuracy) when eNTK features are extracted from a FedAvg-trained model. Also, we find that TCT is stable with respect to the choice of communication rounds $T_1$ in {Stage 1}. For example, models trained by TCT with $T_1 \geq 60$ achieve similar performance.

\vspace{-0.1in}
\paragraph{Effect of normalization.} In Figure~\ref{fig:method-fig-2}, we investigate the role of normalization on TCT by comparing TCT run with normalized and unnormalized eNTK features. The same number of local steps ($M=500$) is applied for both settings. We tune the learning rate $\eta$ for each setting and plot the run that performs best (as measured in training accuracy). 
The results in Figure~\ref{fig:method-fig-2} suggest that the normalization step in TCT significantly improves the communication efficiency by increasing convergence speed. 
In particular, TCT with normalization converges to nearly $100\%$ training accuracy in approximately 40 communication rounds, which is much faster than TCT without normalization. 

\vspace{-0.1in}
\paragraph{Pre-training vs. Bootstrapping.} In Appendix~\ref{subsec:pretraining-experiments}, we explore the effect of starting from a pre-trained model instead of relying on bootstrapping to learn the features. We find that pre-training further improves the performance of TCT and completely erases the gap between centralized and federated learning.

\vspace{-0.3em}
\section{Conclusion}
\vspace{-0.3em}
We have argued that nonconvexity poses a significant challenge for federated learning algorithms. We found that a neural network trained in such a manner does learn useful features, but fails to use them and thus has poor overall accuracy. To sidestep this issue, we proposed a \emph{train-convexify-train} procedure: first, train the neural network using FedAvg; then, optimize (using SCAFFOLD) a convex approximation of the model obtained using its empirical neural tangent kernel. We showed that the first stage extracts meaningful features, whereas the second stage learns to utilize these features to obtain a highly performant model. The resulting algorithm is significantly faster and more stable to hyper-parameters than previous federated learning methods. Finally, we also showed that given a good pre-pretrained feature extractor, our convexify-train procedure fully closes the gap between centralized and federated learning. 

Our algorithm adds to the growing body of work using eNTK to \emph{linearize} neural networks and obtain tractable convex approximations. However, unlike most of these past works which only work with pre-trained models, our bootstrapping allows training models from scratch.
Finally, we stress that the success of our approach underscores the need to revisit theoretical understanding of heterogeneous federated learning. Nonconvexity seems to play an outsized role but its effect in FL has hitherto been unexplored. In particular, black-box notions of difficulty such as gradient dissimilarity or distances between client optima seem insufficient to capture practical performance. It is likely that further progress in the field (e.g. federated pre-training of foundational models), will require tackling the issue of nonconvexity head on.


\bibliographystyle{plainnat}
\bibliography{reference}

\begin{thebibliography}{81}
\providecommand{\natexlab}[1]{#1}
\providecommand{\url}[1]{\texttt{#1}}
\expandafter\ifx\csname urlstyle\endcsname\relax
  \providecommand{\doi}[1]{doi: #1}\else
  \providecommand{\doi}{doi: \begingroup \urlstyle{rm}\Url}\fi

\bibitem[Acar et~al.(2021)Acar, Zhao, Matas, Mattina, Whatmough, and
  Saligrama]{acar2021federated}
Durmus Alp~Emre Acar, Yue Zhao, Ramon Matas, Matthew Mattina, Paul Whatmough,
  and Venkatesh Saligrama.
\newblock Federated learning based on dynamic regularization.
\newblock In \emph{International Conference on Learning Representations}, 2021.
\newblock URL \url{https://openreview.net/forum?id=B7v4QMR6Z9w}.

\bibitem[Achille et~al.(2021)Achille, Golatkar, Ravichandran, Polito, and
  Soatto]{achille2021lqf}
Alessandro Achille, Aditya Golatkar, Avinash Ravichandran, Marzia Polito, and
  Stefano Soatto.
\newblock Lqf: Linear quadratic fine-tuning.
\newblock In \emph{Proceedings of the IEEE/CVF Conference on Computer Vision
  and Pattern Recognition}, pages 15729--15739, 2021.

\bibitem[Afonin and Karimireddy(2021)]{afonin2021towards}
Andrei Afonin and Sai~Praneeth Karimireddy.
\newblock Towards model agnostic federated learning using knowledge
  distillation.
\newblock \emph{arXiv preprint arXiv:2110.15210}, 2021.

\bibitem[Alistarh et~al.(2017)Alistarh, Grubic, Li, Tomioka, and
  Vojnovic]{alistarh2017qsgd}
Dan Alistarh, Demjan Grubic, Jerry Li, Ryota Tomioka, and Milan Vojnovic.
\newblock {QSGD}: Communication-efficient {SGD} via gradient quantization and
  encoding.
\newblock \emph{Advances in Neural Information Processing Systems}, 30, 2017.

\bibitem[Avron et~al.()Avron, Clarkson, and Woodruff]{avron17sharper}
Haim Avron, Kenneth~L. Clarkson, and David~P. Woodruff.
\newblock Sharper bounds for regularized data fitting.
\newblock In \emph{Approximation, Randomization, and Combinatorial
  Optimization. Algorithms and Techniques, {}}.

\bibitem[Bagdasaryan et~al.(2020)Bagdasaryan, Veit, Hua, Estrin, and
  Shmatikov]{bagdasaryan2020backdoor}
Eugene Bagdasaryan, Andreas Veit, Yiqing Hua, Deborah Estrin, and Vitaly
  Shmatikov.
\newblock How to backdoor federated learning.
\newblock In \emph{International Conference on Artificial Intelligence and
  Statistics}, pages 2938--2948. PMLR, 2020.

\bibitem[Blanchard et~al.(2017)Blanchard, El~Mhamdi, Guerraoui, and
  Stainer]{blanchard2017machine}
Peva Blanchard, El~Mahdi El~Mhamdi, Rachid Guerraoui, and Julien Stainer.
\newblock Machine learning with adversaries: Byzantine tolerant gradient
  descent.
\newblock \emph{Advances in Neural Information Processing Systems}, 30, 2017.

\bibitem[Bonawitz et~al.(2021)Bonawitz, Kairouz, McMahan, and
  Ramage]{bonawitz2021federated}
Kallista Bonawitz, Peter Kairouz, Brendan McMahan, and Daniel Ramage.
\newblock Federated learning and privacy: Building privacy-preserving systems
  for machine learning and data science on decentralized data.
\newblock \emph{Queue}, 19\penalty0 (5):\penalty0 87--114, 2021.

\bibitem[Bonawitz et~al.(2017)Bonawitz, Ivanov, Kreuter, Marcedone, McMahan,
  Patel, Ramage, Segal, and Seth]{bonawitz2017practical}
Keith Bonawitz, Vladimir Ivanov, Ben Kreuter, Antonio Marcedone, H~Brendan
  McMahan, Sarvar Patel, Daniel Ramage, Aaron Segal, and Karn Seth.
\newblock Practical secure aggregation for privacy-preserving machine learning.
\newblock In \emph{Proceedings of the 2017 ACM SIGSAC Conference on Computer
  and Communications Security}, pages 1175--1191, 2017.

\bibitem[Bonawitz et~al.(2019)Bonawitz, Eichner, Grieskamp, Huba, Ingerman,
  Ivanov, Kiddon, Kone{\v{c}}n{\`y}, Mazzocchi, McMahan,
  et~al.]{bonawitz2019towards}
Keith Bonawitz, Hubert Eichner, Wolfgang Grieskamp, Dzmitry Huba, Alex
  Ingerman, Vladimir Ivanov, Chloe Kiddon, Jakub Kone{\v{c}}n{\`y}, Stefano
  Mazzocchi, Brendan McMahan, et~al.
\newblock Towards federated learning at scale: System design.
\newblock \emph{Proceedings of Machine Learning and Systems}, 1:\penalty0
  374--388, 2019.

\bibitem[Charles et~al.(2021)Charles, Garrett, Huo, Shmulyian, and
  Smith]{charles2021large}
Zachary Charles, Zachary Garrett, Zhouyuan Huo, Sergei Shmulyian, and Virginia
  Smith.
\newblock On large-cohort training for federated learning.
\newblock \emph{Advances in Neural Information Processing Systems}, 34, 2021.

\bibitem[Chayti and Karimireddy(2022)]{chayti2022optimization}
El~Mahdi Chayti and Sai~Praneeth Karimireddy.
\newblock Optimization with access to auxiliary information.
\newblock \emph{arXiv preprint arXiv:2206.00395}, 2022.

\bibitem[Collins et~al.(2021)Collins, Hassani, Mokhtari, and
  Shakkottai]{collins2021exploiting}
Liam Collins, Hamed Hassani, Aryan Mokhtari, and Sanjay Shakkottai.
\newblock Exploiting shared representations for personalized federated
  learning.
\newblock In \emph{International Conference on Machine Learning}, pages
  2089--2099. PMLR, 2021.

\bibitem[Dean et~al.(2012)Dean, Corrado, Monga, Chen, Devin, Mao, Ranzato,
  Senior, Tucker, Yang, et~al.]{dean2012large}
Jeffrey Dean, Greg Corrado, Rajat Monga, Kai Chen, Matthieu Devin, Mark Mao,
  Marc'aurelio Ranzato, Andrew Senior, Paul Tucker, Ke~Yang, et~al.
\newblock Large scale distributed deep networks.
\newblock \emph{Advances in Neural Information Processing Systems}, 25, 2012.

\bibitem[Defazio et~al.(2014)Defazio, Bach, and
  Lacoste-Julien]{defazio2014saga}
Aaron Defazio, Francis Bach, and Simon Lacoste-Julien.
\newblock Saga: A fast incremental gradient method with support for
  non-strongly convex composite objectives.
\newblock \emph{Advances in Neural Information Processing Systems}, 27, 2014.

\bibitem[Deng et~al.(2020)Deng, Kamani, and Mahdavi]{deng2020adaptive}
Yuyang Deng, Mohammad~Mahdi Kamani, and Mehrdad Mahdavi.
\newblock Adaptive personalized federated learning.
\newblock \emph{arXiv preprint arXiv:2003.13461}, 2020.

\bibitem[du~Terrail et~al.(2022)du~Terrail, Ayed, Cyffers, Grimberg, He, Loeb,
  Mangold, Marchand, Marfoq, Mushtaq, et~al.]{du2022flamby}
Jean~Ogier du~Terrail, Samy-Safwan Ayed, Edwige Cyffers, Felix Grimberg,
  Chaoyang He, Regis Loeb, Paul Mangold, Tanguy Marchand, Othmane Marfoq, Erum
  Mushtaq, et~al.
\newblock Flamby: Datasets and benchmarks for cross-silo federated learning in
  realistic settings.
\newblock 2022.

\bibitem[Fallah et~al.(2020)Fallah, Mokhtari, and
  Ozdaglar]{fallah2020personalized}
Alireza Fallah, Aryan Mokhtari, and Asuman Ozdaglar.
\newblock Personalized federated learning: A meta-learning approach.
\newblock \emph{arXiv preprint arXiv:2002.07948}, 2020.

\bibitem[Fang et~al.(2020)Fang, Cao, Jia, and Gong]{fang2020local}
Minghong Fang, Xiaoyu Cao, Jinyuan Jia, and Neil Gong.
\newblock Local model poisoning attacks to $\{$Byzantine-Robust$\}$ federated
  learning.
\newblock In \emph{29th USENIX Security Symposium (USENIX Security 20)}, pages
  1605--1622, 2020.

\bibitem[Fort et~al.(2020)Fort, Dziugaite, Paul, Kharaghani, Roy, and
  Ganguli]{fortntk}
Stanislav Fort, Gintare~Karolina Dziugaite, Mansheej Paul, Sepideh Kharaghani,
  Daniel~M Roy, and Surya Ganguli.
\newblock Deep learning versus kernel learning: an empirical study of loss
  landscape geometry and the time evolution of the neural tangent kernel.
\newblock In \emph{Advances in Neural Information Processing Systems},
  volume~33, pages 5850--5861, 2020.

\bibitem[Fung et~al.(2018)Fung, Yoon, and Beschastnikh]{fung2018mitigating}
Clement Fung, Chris~JM Yoon, and Ivan Beschastnikh.
\newblock Mitigating sybils in federated learning poisoning.
\newblock \emph{arXiv preprint arXiv:1808.04866}, 2018.

\bibitem[Goldblum et~al.(2019)Goldblum, Geiping, Schwarzschild, Moeller, and
  Goldstein]{goldblum2019truth}
Micah Goldblum, Jonas Geiping, Avi Schwarzschild, Michael Moeller, and Tom
  Goldstein.
\newblock Truth or backpropaganda? an empirical investigation of deep learning
  theory.
\newblock \emph{arXiv preprint arXiv:1910.00359}, 2019.

\bibitem[Goyal et~al.(2017)Goyal, Doll{\'a}r, Girshick, Noordhuis, Wesolowski,
  Kyrola, Tulloch, Jia, and He]{goyal2017accurate}
Priya Goyal, Piotr Doll{\'a}r, Ross Girshick, Pieter Noordhuis, Lukasz
  Wesolowski, Aapo Kyrola, Andrew Tulloch, Yangqing Jia, and Kaiming He.
\newblock Accurate, large minibatch {SGD}: Training {I}magenet in 1 hour.
\newblock \emph{arXiv preprint arXiv:1706.02677}, 2017.

\bibitem[Haddadpour et~al.(2021)Haddadpour, Kamani, Mokhtari, and
  Mahdavi]{haddadpour2021federated}
Farzin Haddadpour, Mohammad~Mahdi Kamani, Aryan Mokhtari, and Mehrdad Mahdavi.
\newblock Federated learning with compression: Unified analysis and sharp
  guarantees.
\newblock In \emph{International Conference on Artificial Intelligence and
  Statistics}, pages 2350--2358. PMLR, 2021.

\bibitem[He et~al.(2016)He, Zhang, Ren, and Sun]{he2016deep}
Kaiming He, Xiangyu Zhang, Shaoqing Ren, and Jian Sun.
\newblock Deep residual learning for image recognition.
\newblock In \emph{Proceedings of the IEEE conference on computer vision and
  pattern recognition}, pages 770--778, 2016.

\bibitem[He et~al.(2022)He, Karimireddy, and Jaggi]{he2022byzantine}
Lie He, Sai~Praneeth Karimireddy, and Martin Jaggi.
\newblock Byzantine-robust decentralized learning via self-centered clipping.
\newblock \emph{arXiv preprint arXiv:2202.01545}, 2022.

\bibitem[Hsieh et~al.(2020)Hsieh, Phanishayee, Mutlu, and
  Gibbons]{hsieh2020non}
Kevin Hsieh, Amar Phanishayee, Onur Mutlu, and Phillip Gibbons.
\newblock The non-iid data quagmire of decentralized machine learning.
\newblock In \emph{International Conference on Machine Learning}, pages
  4387--4398. PMLR, 2020.

\bibitem[Hsu et~al.(2019)Hsu, Qi, and Brown]{hsu2019measuring}
Tzu-Ming~Harry Hsu, Hang Qi, and Matthew Brown.
\newblock Measuring the effects of non-identical data distribution for
  federated visual classification.
\newblock \emph{arXiv preprint arXiv:1909.06335}, 2019.

\bibitem[Hui and Belkin(2020)]{hui2020evaluation}
Like Hui and Mikhail Belkin.
\newblock Evaluation of neural architectures trained with square loss vs
  cross-entropy in classification tasks.
\newblock \emph{arXiv preprint arXiv:2006.07322}, 2020.

\bibitem[Iandola et~al.(2016)Iandola, Moskewicz, Ashraf, and
  Keutzer]{iandola2016firecaffe}
Forrest~N Iandola, Matthew~W Moskewicz, Khalid Ashraf, and Kurt Keutzer.
\newblock Firecaffe: near-linear acceleration of deep neural network training
  on compute clusters.
\newblock In \emph{Proceedings of the IEEE Conference on Computer Vision and
  Pattern Recognition}, pages 2592--2600, 2016.

\bibitem[Ioffe and Szegedy(2015)]{ioffe2015batch}
Sergey Ioffe and Christian Szegedy.
\newblock Batch normalization: Accelerating deep network training by reducing
  internal covariate shift.
\newblock In \emph{International conference on machine learning}, pages
  448--456. PMLR, 2015.

\bibitem[Izmailov et~al.(2018)Izmailov, Podoprikhin, Garipov, Vetrov, and
  Wilson]{izmailov2018averaging}
Pavel Izmailov, Dmitrii Podoprikhin, Timur Garipov, Dmitry Vetrov, and
  Andrew~Gordon Wilson.
\newblock Averaging weights leads to wider optima and better generalization.
\newblock \emph{arXiv preprint arXiv:1803.05407}, 2018.

\bibitem[Jacot et~al.(2018)Jacot, Gabriel, and Hongler]{jacot2018neural}
Arthur Jacot, Franck Gabriel, and Cl{\'e}ment Hongler.
\newblock Neural tangent kernel: Convergence and generalization in neural
  networks.
\newblock \emph{Advances in Neural Information Processing Systems}, 31, 2018.

\bibitem[Johnson and Zhang(2013)]{johnson2013accelerating}
Rie Johnson and Tong Zhang.
\newblock Accelerating stochastic gradient descent using predictive variance
  reduction.
\newblock \emph{Advances in Neural Information Processing Systems}, 26, 2013.

\bibitem[Jones and Tonetti(2020)]{jones2020nonrivalry}
Charles~I Jones and Christopher Tonetti.
\newblock Nonrivalry and the economics of data.
\newblock \emph{American Economic Review}, 110\penalty0 (9):\penalty0 2819--58,
  2020.

\bibitem[Kairouz et~al.(2021)Kairouz, McMahan, et~al.]{kairouz2019federated}
Peter Kairouz, H.~Brendan McMahan, et~al.
\newblock Advances and open problems in federated learning.
\newblock \emph{Foundations and Trends{\textregistered} in Machine Learning},
  14\penalty0 (1--2):\penalty0 1--210, 2021.

\bibitem[Karimireddy et~al.(2020{\natexlab{a}})Karimireddy, Jaggi, Kale, Mohri,
  Reddi, Stich, and Suresh]{karimireddy2020mime}
Sai~Praneeth Karimireddy, Martin Jaggi, Satyen Kale, Mehryar Mohri, Sashank~J
  Reddi, Sebastian~U Stich, and Ananda~Theertha Suresh.
\newblock Mime: Mimicking centralized stochastic algorithms in federated
  learning.
\newblock \emph{arXiv preprint arXiv:2008.03606}, 2020{\natexlab{a}}.

\bibitem[Karimireddy et~al.(2020{\natexlab{b}})Karimireddy, Kale, Mohri, Reddi,
  Stich, and Suresh]{karimireddy2020scaffold}
Sai~Praneeth Karimireddy, Satyen Kale, Mehryar Mohri, Sashank Reddi, Sebastian
  Stich, and Ananda~Theertha Suresh.
\newblock Scaffold: Stochastic controlled averaging for federated learning.
\newblock In \emph{International Conference on Machine Learning}, pages
  5132--5143. PMLR, 2020{\natexlab{b}}.

\bibitem[Karimireddy et~al.(2021)Karimireddy, He, and
  Jaggi]{karimireddy2022byzantine}
Sai~Praneeth Karimireddy, Lie He, and Martin Jaggi.
\newblock Byzantine-robust learning on heterogeneous datasets via bucketing.
\newblock In \emph{International Conference on Learning Representations}, 2021.

\bibitem[Krizhevsky et~al.(2009)Krizhevsky, Hinton,
  et~al.]{krizhevsky2009learning}
Alex Krizhevsky, Geoffrey Hinton, et~al.
\newblock Learning multiple layers of features from tiny images.
\newblock 2009.

\bibitem[Kulkarni et~al.(2020)Kulkarni, Kulkarni, and Pant]{kulkarni2020survey}
Viraj Kulkarni, Milind Kulkarni, and Aniruddha Pant.
\newblock Survey of personalization techniques for federated learning.
\newblock In \emph{2020 Fourth World Conference on Smart Trends in Systems,
  Security and Sustainability (WorldS4)}, pages 794--797. IEEE, 2020.

\bibitem[Kulynych et~al.(2020)Kulynych, Madras, Milli, Raji, Zhou, and
  Zemel]{paml2020}
Bogdan Kulynych, David Madras, Smitha Milli, Inioluwa~Deborah Raji, Angela
  Zhou, and Richard Zemel.
\newblock Participatory approaches to machine learning.
\newblock International Conference on Machine Learning Workshop, 2020.

\bibitem[Lee et~al.(2019)Lee, Xiao, Schoenholz, Bahri, Novak, Sohl-Dickstein,
  and Pennington]{lee2019wide}
Jaehoon Lee, Lechao Xiao, Samuel Schoenholz, Yasaman Bahri, Roman Novak, Jascha
  Sohl-Dickstein, and Jeffrey Pennington.
\newblock Wide neural networks of any depth evolve as linear models under
  gradient descent.
\newblock \emph{Advances in Neural Information Processing Systems}, 32, 2019.

\bibitem[Li et~al.(2021{\natexlab{a}})Li, Diao, Chen, and He]{li2021federated}
Qinbin Li, Yiqun Diao, Quan Chen, and Bingsheng He.
\newblock Federated learning on non-iid data silos: An experimental study.
\newblock \emph{arXiv preprint arXiv:2102.02079}, 2021{\natexlab{a}}.

\bibitem[Li et~al.(2021{\natexlab{b}})Li, He, and Song]{li2021model}
Qinbin Li, Bingsheng He, and Dawn Song.
\newblock Model-contrastive federated learning.
\newblock In \emph{Proceedings of the IEEE/CVF Conference on Computer Vision
  and Pattern Recognition}, pages 10713--10722, 2021{\natexlab{b}}.

\bibitem[Li et~al.(2019)Li, Sanjabi, Beirami, and Smith]{li2019fair}
Tian Li, Maziar Sanjabi, Ahmad Beirami, and Virginia Smith.
\newblock Fair resource allocation in federated learning.
\newblock \emph{arXiv preprint arXiv:1905.10497}, 2019.

\bibitem[Li et~al.(2020)Li, Sahu, Zaheer, Sanjabi, Talwalkar, and
  Smith]{li2020federated}
Tian Li, Anit~Kumar Sahu, Manzil Zaheer, Maziar Sanjabi, Ameet Talwalkar, and
  Virginia Smith.
\newblock Federated optimization in heterogeneous networks.
\newblock \emph{Proceedings of Machine Learning and Systems}, 2:\penalty0
  429--450, 2020.

\bibitem[Lin et~al.(2020)Lin, Kong, Stich, and Jaggi]{lin2020ensemble}
Tao Lin, Lingjing Kong, Sebastian~U Stich, and Martin Jaggi.
\newblock Ensemble distillation for robust model fusion in federated learning.
\newblock \emph{Advances in Neural Information Processing Systems},
  33:\penalty0 2351--2363, 2020.

\bibitem[Lin et~al.(2021)Lin, Karimireddy, Stich, and Jaggi]{lin2021quasi}
Tao Lin, Sai~Praneeth Karimireddy, Sebastian~U Stich, and Martin Jaggi.
\newblock Quasi-global momentum: Accelerating decentralized deep learning on
  heterogeneous data.
\newblock \emph{arXiv preprint arXiv:2102.04761}, 2021.

\bibitem[Long(2021)]{long2021properties}
Philip~M Long.
\newblock Properties of the after kernel.
\newblock \emph{arXiv preprint arXiv:2105.10585}, 2021.

\bibitem[Lyu et~al.(2020)Lyu, Yu, and Yang]{lyu2020threats}
Lingjuan Lyu, Han Yu, and Qiang Yang.
\newblock Threats to federated learning: A survey.
\newblock \emph{arXiv preprint arXiv:2003.02133}, 2020.

\bibitem[McMahan et~al.(2017)McMahan, Moore, Ramage, Hampson, and
  y~Arcas]{mcmahan2017communication}
Brendan McMahan, Eider Moore, Daniel Ramage, Seth Hampson, and Blaise~Aguera
  y~Arcas.
\newblock Communication-efficient learning of deep networks from decentralized
  data.
\newblock In \emph{Artificial Intelligence and Statistics}, pages 1273--1282.
  PMLR, 2017.

\bibitem[Mishchenko et~al.(2022)Mishchenko, Malinovsky, Stich, and
  Richt{\'a}rik]{mishchenko2022proxskip}
Konstantin Mishchenko, Grigory Malinovsky, Sebastian Stich, and Peter
  Richt{\'a}rik.
\newblock Proxskip: Yes! local gradient steps provably lead to communication
  acceleration! finally!
\newblock \emph{arXiv preprint arXiv:2202.09357}, 2022.

\bibitem[Mohri et~al.(2019)Mohri, Sivek, and Suresh]{mohri2019agnostic}
Mehryar Mohri, Gary Sivek, and Ananda~Theertha Suresh.
\newblock Agnostic federated learning.
\newblock In \emph{International Conference on Machine Learning}, pages
  4615--4625. PMLR, 2019.

\bibitem[Mothukuri et~al.(2021)Mothukuri, Parizi, Pouriyeh, Huang,
  Dehghantanha, and Srivastava]{mothukuri2021survey}
Viraaji Mothukuri, Reza~M Parizi, Seyedamin Pouriyeh, Yan Huang, Ali
  Dehghantanha, and Gautam Srivastava.
\newblock A survey on security and privacy of federated learning.
\newblock \emph{Future Generation Computer Systems}, 115:\penalty0 619--640,
  2021.

\bibitem[Mu et~al.(2020)Mu, Liang, and Li]{mu2020gradients}
Fangzhou Mu, Yingyu Liang, and Yin Li.
\newblock Gradients as features for deep representation learning.
\newblock \emph{arXiv preprint arXiv:2004.05529}, 2020.

\bibitem[Ozkara et~al.(2021)Ozkara, Singh, Data, and Diggavi]{ozkara2021quped}
Kaan Ozkara, Navjot Singh, Deepesh Data, and Suhas Diggavi.
\newblock Quped: Quantized personalization via distillation with applications
  to federated learning.
\newblock \emph{Advances in Neural Information Processing Systems}, 34, 2021.

\bibitem[Pentland et~al.(2021)Pentland, Lipton, and
  Hardjono]{pentland2021building}
Alex Pentland, Alexander Lipton, and Thomas Hardjono.
\newblock \emph{Building the New Economy: Data as Capital}.
\newblock MIT Press, 2021.

\bibitem[Ramaswamy et~al.(2020)Ramaswamy, Thakkar, Mathews, Andrew, McMahan,
  and Beaufays]{ramaswamy2020training}
Swaroop Ramaswamy, Om~Thakkar, Rajiv Mathews, Galen Andrew, H~Brendan McMahan,
  and Fran{\c{c}}oise Beaufays.
\newblock Training production language models without memorizing user data.
\newblock \emph{arXiv preprint arXiv:2009.10031}, 2020.

\bibitem[Reddi et~al.(2021)Reddi, Charles, Zaheer, Garrett, Rush,
  Kone{\v{c}}n{\'y}, Kumar, and McMahan]{reddi2021adaptive}
Sashank~J. Reddi, Zachary Charles, Manzil Zaheer, Zachary Garrett, Keith Rush,
  Jakub Kone{\v{c}}n{\'y}, Sanjiv Kumar, and Hugh~Brendan McMahan.
\newblock Adaptive federated optimization.
\newblock In \emph{International Conference on Learning Representations}, 2021.
\newblock URL \url{https://openreview.net/forum?id=LkFG3lB13U5}.

\bibitem[Shi et~al.(2021)Shi, Yu, and Leung]{shi2021survey}
Yuxin Shi, Han Yu, and Cyril Leung.
\newblock A survey of fairness-aware federated learning.
\newblock \emph{arXiv preprint arXiv:2111.01872}, 2021.

\bibitem[Singh and Jaggi(2020)]{singh2020model}
Sidak~Pal Singh and Martin Jaggi.
\newblock Model fusion via optimal transport.
\newblock \emph{Advances in Neural Information Processing Systems},
  33:\penalty0 22045--22055, 2020.

\bibitem[So et~al.(2020)So, G{\"u}ler, and Avestimehr]{so2020byzantine}
Jinhyun So, Ba{\c{s}}ak G{\"u}ler, and A~Salman Avestimehr.
\newblock Byzantine-resilient secure federated learning.
\newblock \emph{IEEE Journal on Selected Areas in Communications}, 39\penalty0
  (7):\penalty0 2168--2181, 2020.

\bibitem[Stich and Karimireddy(2020)]{stich2020error}
Sebastian~U Stich and Sai~Praneeth Karimireddy.
\newblock The error-feedback framework: Better rates for sgd with delayed
  gradients and compressed updates.
\newblock \emph{Journal of Machine Learning Research}, 21:\penalty0 1--36,
  2020.

\bibitem[Sun et~al.(2019)Sun, Kairouz, Suresh, and McMahan]{sun2019can}
Ziteng Sun, Peter Kairouz, Ananda~Theertha Suresh, and H~Brendan McMahan.
\newblock Can you really backdoor federated learning?
\newblock \emph{arXiv preprint arXiv:1911.07963}, 2019.

\bibitem[Suresh et~al.(2017)Suresh, Felix, Kumar, and
  McMahan]{suresh2017distributed}
Ananda~Theertha Suresh, X~Yu Felix, Sanjiv Kumar, and H~Brendan McMahan.
\newblock Distributed mean estimation with limited communication.
\newblock In \emph{International Conference on Machine Learning}, pages
  3329--3337. PMLR, 2017.

\bibitem[Tan et~al.(2021)Tan, Long, Liu, Zhou, Lu, Jiang, and
  Zhang]{tan2021fedproto}
Yue Tan, Guodong Long, Lu~Liu, Tianyi Zhou, Qinghua Lu, Jing Jiang, and Chengqi
  Zhang.
\newblock Fedproto: Federated prototype learning over heterogeneous devices.
\newblock \emph{arXiv preprint arXiv:2105.00243}, 2021.

\bibitem[Wang et~al.(2020{\natexlab{a}})Wang, Sreenivasan, Rajput, Vishwakarma,
  Agarwal, Sohn, Lee, and Papailiopoulos]{wang2020attack}
Hongyi Wang, Kartik Sreenivasan, Shashank Rajput, Harit Vishwakarma, Saurabh
  Agarwal, Jy-yong Sohn, Kangwook Lee, and Dimitris Papailiopoulos.
\newblock Attack of the tails: Yes, you really can backdoor federated learning.
\newblock \emph{Advances in Neural Information Processing Systems},
  33:\penalty0 16070--16084, 2020{\natexlab{a}}.

\bibitem[Wang et~al.(2019{\natexlab{a}})Wang, Tantia, Ballas, and
  Rabbat]{wang2019slowmo}
Jianyu Wang, Vinayak Tantia, Nicolas Ballas, and Michael Rabbat.
\newblock Slowmo: Improving communication-efficient distributed sgd with slow
  momentum.
\newblock \emph{arXiv preprint arXiv:1910.00643}, 2019{\natexlab{a}}.

\bibitem[Wang et~al.(2020{\natexlab{b}})Wang, Liu, Liang, Joshi, and
  Poor]{wang2020tackling}
Jianyu Wang, Qinghua Liu, Hao Liang, Gauri Joshi, and H~Vincent Poor.
\newblock Tackling the objective inconsistency problem in heterogeneous
  federated optimization.
\newblock \emph{Advances in Neural Information Processing Systems},
  33:\penalty0 7611--7623, 2020{\natexlab{b}}.

\bibitem[Wang et~al.(2021)Wang, Charles, Xu, Joshi, McMahan, Al-Shedivat,
  Andrew, Avestimehr, Daly, Data, et~al.]{wang2021field}
Jianyu Wang, Zachary Charles, Zheng Xu, Gauri Joshi, H~Brendan McMahan, Maruan
  Al-Shedivat, Galen Andrew, Salman Avestimehr, Katharine Daly, Deepesh Data,
  et~al.
\newblock A field guide to federated optimization.
\newblock \emph{arXiv preprint arXiv:2107.06917}, 2021.

\bibitem[Wang et~al.(2022)Wang, Das, Joshi, Kale, Xu, and
  Zhang]{wang2022unreasonable}
Jianyu Wang, Rudrajit Das, Gauri Joshi, Satyen Kale, Zheng Xu, and Tong Zhang.
\newblock On the unreasonable effectiveness of federated averaging with
  heterogeneous data.
\newblock \emph{arXiv preprint arXiv:2206.04723}, 2022.

\bibitem[Wang et~al.(2019{\natexlab{b}})Wang, Tuor, Salonidis, Leung, Makaya,
  He, and Chan]{wang2019adaptive}
Shiqiang Wang, Tiffany Tuor, Theodoros Salonidis, Kin~K Leung, Christian
  Makaya, Ting He, and Kevin Chan.
\newblock Adaptive federated learning in resource constrained edge computing
  systems.
\newblock \emph{IEEE Journal on Selected Areas in Communications}, 37\penalty0
  (6):\penalty0 1205--1221, 2019{\natexlab{b}}.

\bibitem[Wei et~al.(2022)Wei, Hu, and Steinhardt]{wei2022more}
Alexander Wei, Wei Hu, and Jacob Steinhardt.
\newblock More than a toy: Random matrix models predict how real-world neural
  representations generalize.
\newblock \emph{arXiv preprint arXiv:2203.06176}, 2022.

\bibitem[Woodworth et~al.(2020)Woodworth, Patel, and
  Srebro]{woodworth2020minibatch}
Blake~E Woodworth, Kumar~Kshitij Patel, and Nati Srebro.
\newblock Minibatch vs local sgd for heterogeneous distributed learning.
\newblock \emph{Advances in Neural Information Processing Systems},
  33:\penalty0 6281--6292, 2020.

\bibitem[Wortsman et~al.(2022)Wortsman, Ilharco, Gadre, Roelofs, Gontijo-Lopes,
  Morcos, Namkoong, Farhadi, Carmon, Kornblith, et~al.]{wortsman2022model}
Mitchell Wortsman, Gabriel Ilharco, Samir~Yitzhak Gadre, Rebecca Roelofs,
  Raphael Gontijo-Lopes, Ari~S Morcos, Hongseok Namkoong, Ali Farhadi, Yair
  Carmon, Simon Kornblith, et~al.
\newblock Model soups: averaging weights of multiple fine-tuned models improves
  accuracy without increasing inference time.
\newblock \emph{arXiv preprint arXiv:2203.05482}, 2022.

\bibitem[Wu et~al.(2020)Wu, He, and Chen]{wu2020personalized}
Qiong Wu, Kaiwen He, and Xu~Chen.
\newblock Personalized federated learning for intelligent {IoT} applications: A
  cloud-edge based framework.
\newblock \emph{IEEE Open Journal of the Computer Society}, 1:\penalty0 35--44,
  2020.

\bibitem[Wu and He(2018)]{wu2018group}
Yuxin Wu and Kaiming He.
\newblock Group normalization.
\newblock In \emph{Proceedings of the European Conference on Computer Vision
  (ECCV)}, pages 3--19, 2018.

\bibitem[Xiao et~al.(2017)Xiao, Rasul, and Vollgraf]{xiao2017fashion}
Han Xiao, Kashif Rasul, and Roland Vollgraf.
\newblock Fashion-{MNIST}: a novel image dataset for benchmarking machine
  learning algorithms.
\newblock \emph{arXiv preprint arXiv:1708.07747}, 2017.

\bibitem[Yu et~al.(2021)Yu, Zhang, Qin, Xu, Wang, Liu, Tian, and
  Chen]{yu2021fed2}
Fuxun Yu, Weishan Zhang, Zhuwei Qin, Zirui Xu, Di~Wang, Chenchen Liu, Zhi Tian,
  and Xiang Chen.
\newblock Fed2: Feature-aligned federated learning.
\newblock In \emph{Proceedings of the 27th ACM SIGKDD Conference on Knowledge
  Discovery \& Data Mining}, pages 2066--2074, 2021.

\bibitem[Zancato et~al.(2020)Zancato, Achille, Ravichandran, Bhotika, and
  Soatto]{zancato2020predicting}
Luca Zancato, Alessandro Achille, Avinash Ravichandran, Rahul Bhotika, and
  Stefano Soatto.
\newblock Predicting training time without training.
\newblock \emph{Advances in Neural Information Processing Systems},
  33:\penalty0 6136--6146, 2020.

\end{thebibliography}

\newpage
\appendix

\begin{center}
    \textbf{\Large Appendix}
\end{center}
\vspace{0.1in}
\section{Additional Details About Our Algorithm}\label{sec:appendix-algorithm-details}

\subsection{An Efficient Implementation of SCAFFOLD}

\begin{algorithm}[!h]
\begin{algorithmic}
\STATE \textbf{Input:} losses $\{L_k\}$, $k\in [K]$. Number of local steps $M$, server model $\theta^0$, learning rate $\eta$.
\STATE \textbf{Initialization:} \quad client corrections $\{h_k^{-1} = \mathbf{0}\}$, local models$\{\widehat{\theta}^{0}_{i}\} = \theta^0, \,\, k \in [K]$
\FOR{round $t=0,1,\ldots, T$} 
\FOR{clients $k=1,\ldots, K$ in parallel} 
\STATE Receive $\theta^{t}$ from server. Update correction
\begin{equation}\label{eq:2}
    \begin{aligned}
    h_{k}^{t} &= h_k^{t-1} + \tfrac{1}{M \eta}(\theta^{t} - \widehat{\theta}^{t}_{k}).
    \end{aligned}
\end{equation}
\STATE Initialize client local model $\widehat\theta_i^{t,0} = \theta^t$.
    \FOR{$m = 1, \dots, M$}
        \STATE Update with a stochastic gradient sampled from local client data
        \begin{equation}\label{eq:appendix-scaffold-sgd-update}
            \begin{aligned}
            \widehat\theta_k^{t, m+1} &= \widehat\theta_k^{t, m} - \eta \left(\nabla L_k(\theta_i^{t, m} ; \xi_k^{t,m}) - h_k^t \right).
            \end{aligned}
        \end{equation}
    \ENDFOR
    \STATE Set $\widehat\theta_k^{t+1} = \widehat\theta_k^{t, M+1}$. Communicate $\widehat\theta_k^{t+1}$ to server.
\ENDFOR
\STATE Aggregate $\theta^{t+1} = \frac{1}{K}\sum_{k=1}^{K}\widehat{\theta}^{t+1}_{k}\,.$\footnotemark
\ENDFOR
\end{algorithmic} \caption{Efficient implementation of SCAFFOLD}\label{Algorithm:Scaffold-eNTK}
\end{algorithm} 
\footnotetext{Note that when different clients have different number of data points, the actual aggregation step is $\theta^{t+1} = \sum_{k=1}^{K} (\nicefrac{n_k}{ \sum_j n_j}) \widehat{\theta}^{t+1}_{k}$. However, we present the simplified version with equal weights for all clients to ease the comparison with the pseudocode in \citet{karimireddy2020scaffold}.}

We describe a more communication efficient implementation of SCAFFOLD which is equivalent to Option II of SCAFFOLD from \cite{karimireddy2020scaffold}. Our implementation only requires a single model to be communicated between the client and server each round, making its communication complexity exactly equivalent to that of FedAvg. To see the equivalence, we prove that our implementation satisfies the following condition for any time step $t \geq 0$:
\begin{align*}
    c_k^{t+1} &:= \frac{1}{M}\sum_{m \in [M]}\nabla L_k(\theta_k^{t, m} ; \xi_k^{t,m})\,, \text{ and}\\
    c^{t+1} &:= \frac{1}{K}\sum_{k\in[K]}c_k^t\,, \text{ we maintain the invariant that}\\
    h_k^{t+1} &= c_k^{t+1} - c^{t+1}\,.
\end{align*}
To see this, note that the local client model after updating in round $t$ is
\begin{align*}
    \widehat\theta_k^{t+1} &= \widehat\theta_k^{t, M+1}\\
    &= \theta^t - \eta\sum_{k \in [K]}\nabla L_k(\theta_k^{t, m} ; \xi_k^{t,m}) - h_k^{t} \\
    &= \theta^t - M \eta (c_k^{t+1} - h_k^t)\,.
\end{align*}
By averaging this over the clients, we can see that the server model is
\[
\theta^{t+1} = \theta^t - M\eta \Big(c^{t+1} - \frac{1}{K}\sum_{l\in [K]} h_l^t\Big).
\]
By induction, suppose that $h_k^{t} = c_k^{t} - c^{t}$. This implies that summing over the clients, it becomes zero; i.e., $\sum_{l\in [K]} h_l^t = 0$. Plugging this and the previous computations, we have
\begin{align*}
    h_{k}^{t+1} &= h_k^{t} + \tfrac{1}{M \eta}(\theta^{t+1} - \widehat{\theta}^{t+1}_{k})\\
    &= h_k^{t} + \tfrac{1}{M \eta}(- M \eta c^{t+1}  + M \eta (c_k^{t+1} - h_k^t))\\
    &= c_k^{t+1} - c^{t+1}\,.
\end{align*}
For the base step at $t=0$, note that $h_i^0 = \bm{0}$. This completes the proof by induction.

\subsection{Additional Implementation Details}

\begin{algorithm}[!h]
\begin{algorithmic}
\STATE \textbf{Input:} input dim $D$, output dim $C$, loss $\ell(\cdot, \cdot): \mathbb{R}^{C \times C} \rightarrow \mathbb{R}$, aggr. weights $\{w_1, \dots, w_K\}$,
\STATE {model $f$ with parameters $\theta \in \mathbb{R}^P$:} $f(x; \theta) = \phi \circ \omega~(x): \mathbb{R}^D \rightarrow \mathbb{R}^C$ (e.g., ResNet18), composed of a feature extractor~$\phi: \mathbb{R}^D \rightarrow \mathbb{R}^E$ and final linear layer $\omega: \mathbb{R}^E \rightarrow \mathbb{R}^C$.
\STATE \textbf{Hyper-parameters:} Local steps $M$ (default 500), Stage-1 lr $\eta_1$ (default 0.01), Stage-1 rounds $T_1$ (default 100), Stage-2 lr $\eta_{2}$ (default $5\cdot 10^{-5}$), Stage-2 rounds $T_2$ (default 100). 
\vspace{1em}
\STATE \textbf{Stage 1 (Bootstrapping):} 
\STATE Initialize server model $\theta^0$.
\FOR{round $t=0,1,\ldots, T_1$} 
\FOR{clients $k=1,\ldots, K$ in parallel} 
\STATE Receive $\theta^{t}$ from server and initialize client local model $\widehat\theta_k^{t,0} = \theta^t$.
    \FOR{$m = 1, \dots, M$}
        \STATE Update with a mini-batch gradient sampled from local client data $(x_k^{t,m}, y_k^{t,m})$
        \begin{equation*}
            \begin{aligned}
            \widehat\theta_k^{t, m+1} &= \widehat\theta_k^{t, m} - \eta_i \left(\nabla_\theta \ell(f(x_k^{t,m} ; \theta_k^{t, m}), y_k^{t,m}) - h_k^t \right).
            \end{aligned}
        \end{equation*}
    \ENDFOR
    \STATE Communicate $\widehat\theta_k^{t+1}$ to server.
\ENDFOR
\STATE Aggregate $\theta^{t+1} = \frac{1}{\sum_k w_k}\sum_{k=1}^{K}w_k\widehat{\theta}^{t+1}_{k}\,.$
\ENDFOR
\vspace{1em}
\STATE \textbf{Stage 2 (Convexification):} 
\STATE Input: Bootstrapped parameters $\theta^B$ decomposed as $\theta^B = \phi^B \circ \omega^B$.
\STATE Randomly re-initialize using fixed seed linear layer $\omega^r$ and define $\theta^0 := \phi^B \circ \omega^r$.
\STATE [Comment:] \emph{Define basis vector $\mathbf{e_1} := (1, 0,\dots, 0)$. For input $x$, $( \mathbf{e}_1^\top f(x; \theta^0))$ is the first logit.}
\STATE Optionally, compute a random sub-sampling mask $\mathscr{S}(\phi)$ over feature params using fixed seed .
\STATE [Comment:] \emph{
For a given input $x$, we will learn parameters $(\varphi, b)$ for prediction as} $$\hat y = \varphi ^\top \phi_\text{eNTK}(x)   + b, \quad \text{where} \quad \phi_\text{eNTK}(x) := \mathscr{S}\left(\nabla_{\phi} ( \mathbf{e}_1^\top f(x; \phi^B \circ \omega^r)) \right).$$
\STATE Compute normalized eNTK features $\tilde\phi_\text{eNTK}(x)$ (mean 0 and variance 1) across clients. 
\STATE Also normalize targets to mean 0 using $\tilde y := y -\frac{1}{C}\mathbf{1}$.
\STATE Run SCAFFOLD (Algorithm~\ref{Algorithm:Scaffold-eNTK}) over params $\psi := (\varphi, b)$ with learning rate $\eta_2$, local steps $M$, initial server params: $\psi^0 = \mathbf{0}$, and client losses $\{L_k\}$ defined over the local data as
\[
    L_k(\psi) := \sum_{(x_k, y_k)} \left(\varphi ^\top \tilde\phi_\text{eNTK}(x_k)   + b  - \tilde y_k\right)^2\,.
\]

\end{algorithmic} \caption{TCT: complete pseudo-code}\label{Algorithm:Boo-NTK-Full-details}
\end{algorithm}

\begin{algorithm}[t]
\caption{Compute eNTK Pseudocode, PyTorch-like}
\label{alg:code}
\definecolor{codeblue}{rgb}{0.25,0.5,0.5}
\definecolor{codekw}{rgb}{0.85, 0.18, 0.50}
\lstset{
  backgroundcolor=\color{white},
  basicstyle=\fontsize{7.5pt}{7.5pt}\ttfamily\selectfont,
  columns=fullflexible,
  breaklines=true,
  captionpos=b,
  commentstyle=\fontsize{7.5pt}{7.5pt}\color{codeblue},
  keywordstyle=\fontsize{7.5pt}{7.5pt}\color{codekw},
}
\begin{lstlisting}[language=python]
def compute_eNTK(model, X, num_params, subsample_size=100000, seed=123):
    """compute eNTK of input X with model"""
    # model: model for linearization
    # X: (n x d), n -- number of samples, d -- input dimension
    # subsample_size: parameter of subsampling operation
    # seed: random seed for subsampling operation
    # num_params: total number of parameters for model
    model.eval()
    params = list(model.parameters())
    torch.manual_seed(seed)
    torch.cuda.manual_seed(seed)
    random_index = torch.randperm(num_params)[:subsample_size]
    eNTKs = torch.zeros((X.size()[0], subsample_size))
    for i in range(X.size()[0]):
        # compute eNTK for the i-th input
        model.zero_grad()
        model.forward(X[i:i+1])[0].backward()
        eNTK = []
        for param in params:
            if param.requires_grad:
                eNTK.append(param.grad.flatten())
        eNTK = torch.cat(eNTK)
        # subsampling
        eNTKs[i, :] = eNTK[random_index]
    return eNTKs
\end{lstlisting}
\end{algorithm}

\paragraph{Additional details about linear regression in TCT.} In our experiments, we normalize the one-hot encoded label of each sample so that the normalized one-hot encoded label has mean 0. More specifically, we subtract $[1/C, \dots, 1/C]^{\top}\in\mathbb{R}^{C\times 1}$ from the one-hot encoding label vector, where $C$ is the number of classes. Further, ~\citet{hui2020evaluation} show that performance for large number of classes can be improved by increasing the penalty for mis-classification and scaling the target from 1 to a larger value (e.g., 30). \citet{achille2021lqf} show that using Leaky-ReLU, and using K-FAC preconditioning further improves the performance. However, we do not explore such optimizations in this work--these (and other optimization tricks for least-squares regression) can be easily incorporated into our framework.

\paragraph{Local learning rate for TCT.} From our experiments, we find that small local learning rates ($\eta\leq 10^{-4}$) achieve good train/test accuracy performance for TCT with the normalization step. When the normalization step in TCT is applied, larger local learning rates diverge. Meanwhile, local learning rates from $[10^{-6}, 10^{-4}]$ achieve similar performance for TCT (as shown in Table~\ref{tab:compare_locallr_table}). On the other hand, without the normalization step, TCT with large learning rate ($\eta\in[0.01, 0.5]$) does not diverge. When running more communication rounds, TCT (without the normalization step) with large learning rate achieves similar performance as the default TCT (with the normalization step). 

\paragraph{Additional details about Stage 2 of TCT.} To solve the linear regression problem in TCT-\textbf{Stage 2}, we use the full batch gradient in Eq.~\eqref{eq:appendix-scaffold-sgd-update} of Algorithm~\ref{Algorithm:Scaffold-eNTK} in our implementation.

\paragraph{Additional details about Figure~\ref{fig:method-fig}.} We consider CIFAR10-[\texttt{\#C=2}] in Figure~\ref{fig:method-fig-1} and \ref{fig:method-fig-2}.

\paragraph{Details about the total amount of compute.} We use NVIDIA 2080 Ti, A4000, and A100 GPUs, and our experiments required around 500 hours of GPU time.

\clearpage
\section{Additional Experimental Results}\label{sec:appendix-additional-exp}

\subsection{Additional Baselines}\label{subsec:appendix-more-baselines}

\textbf{In comparison with FedAdam and FedDyn.} We compare TCT to FedAdam~\citep{reddi2021adaptive} and FedDyn~\citep{acar2021federated} in Table~\ref{tab:table_more_baseline_appendix}. We consider four settings in Table~\ref{tab:table_more_baseline_appendix}, including CIFAR10 ($\texttt{\#C}=2$), CIFAR10 ($\alpha=0.1$),   CIFAR100 ($\alpha=0.001$), and CIFAR100 ($\alpha=0.01$). 
For FedDyn, we perform similar hyperparameter selection as FedAvg; i.e., select local learning rate from $\{0.1, 0.01, 0.001\}$. For FedAdam, following recommendation by~\citep{reddi2021adaptive}, we set the global learning rate as $\eta_{\mathrm{global}}=0.1$ and select local learning rate from $\{10^{-1}, 10^{-1.5}, 10^{-2}, 10^{-2.5}, 10^{-3}\}$.
Similar to results in Table~\ref{tab:main_table}, we find that TCT significantly outperforms the existing methods in high data heterogeneity settings.

\begin{table*}[ht!]
	\centering
	\caption{{The top-1 test accuracy ($\%$) of our algorithm (TCT) vs.\ other federated learning algorithms (FedAdam~\citep{reddi2021adaptive} and FedDyn~\citep{acar2021federated}) evaluated on CIFAR10 and CIFAR100. We vary the degree of data heterogeneity by controlling the $\alpha$ parameter of the symmetric Dirichlet distribution $\mathrm{Dir}_K(\alpha)$ and the $\texttt{\#C}$ parameter for assigning how many labels each client owns. Higher accuracy is better. The highest top-1 accuracy in each setting is highlighted in \textbf{bold}. } 
	}
	\vspace{0.75em}
	\label{tab:table_more_baseline_appendix}
	\resizebox{0.99\textwidth}{!}{
	    \footnotesize
		\begin{tabular}{cccccc}
			\toprule
			\multirow{1}{*}{\bf Methods} & \multicolumn{5}{c}{\bf Datasets}
			  \\
			  \midrule
		\midrule
		   &   CIFAR10 ($\texttt{\#C}=2$) & CIFAR10 ($\alpha=0.1$)  & & CIFAR100 ($\alpha=0.001$) &  CIFAR100 ($\alpha=0.01$)   \\ 
			\cmidrule(lr){2-4}
			\cmidrule(lr){5-6}
			  FedAdam~\citep{reddi2021adaptive}  &  33.52\%   &  62.57\%  & & 30.85\% &  37.16\%    \\ 
			\tablewhitespace
			  FedDyn~\citep{acar2021federated}  &  51.67\%   &  81.03\%  & & 50.86\% &  53.79\%    \\ 
			\tablewhitespace
			  {TCT}  &   \textbf{83.02\%}   &  \textbf{89.21\%}  & & \textbf{69.07\%} &  \textbf{69.66\%}    \\ 
			\bottomrule
		\end{tabular}
	}
	\vspace{-.75em}
\end{table*}

\subsection{Results of Other Architectures }\label{subsec:appendix-other-arch}
In Section~\ref{sec:exp}, we use batch normalization~\citep{ioffe2015batch} as the default normalization layer on CIFAR10 and CIFAR100 datasets, and we denote the ResNet-18 with batch normalization layers by ResNet-18-BN. In Table~\ref{tab:compare_gn_table}, we consider group normalization~\citep{wu2018group} on CIFAR10 and CIFAR100 and let ResNet-18-GN denote the ResNet-18 with group normalization. 
We set \texttt{num\_groups=2} in group normalization layers. As shown in Table~\ref{tab:compare_gn_table}, TCT achieves better performance than FedAvg with ResNet-18-GN on both CIFAR10 and CIFAR100 datasets.  Our experiments indicate that in extremely heterogeneous settings, group norm is insufficient to fix FedAvg.

\begin{table*}[ht]
	\centering
	\caption{{The top-1 test accuracy ($\%$) of our algorithm (TCT) vs.\ FedAvg(-GN) evaluated on  CIFAR10 and CIFAR100. We vary the degree of data heterogeneity by controlling the $\alpha$ parameter of the symmetric Dirichlet distribution $\mathrm{Dir}_K(\alpha)$ and the $\texttt{\#C}$ parameter for assigning how many labels each client owns. Higher accuracy is better. The highest top-1 accuracy in each setting is highlighted in \textbf{bold}. } 
	}
	\vspace{0.75em}
	\label{tab:compare_gn_table}
	    \footnotesize
		\begin{tabular}{ccccccccc}
			\toprule
			\multirow{1}{*}{\bf Datasets} & \multirow{1}{*}{\bf Architectures} &   \multirow{1}{*}{\bf Methods}                 & \multicolumn{4}{c}{\bf Non-i.i.d. degree}   
			  \\
            \midrule
		\midrule
		\multirow{7}{*}{CIFAR-10}   &  &     & $\texttt{\#C}=1$ & $\texttt{\#C}=2$  & $\alpha=0.1$ & 	$\alpha=0.5$
			\\ 
			\cmidrule(lr){1-7} 
			  &  ResNet-18-GN & {FedAvg} &       21.23\%  &  56.80\%  &  84.72\%  &  89.03\%      
			  \\
			  \tablewhitespace 
			  &  ResNet-18-BN & {FedAvg} & 11.27\%  & 56.86\%  & 82.60\%      & 90.43\%          \\
			  \tablewhitespace
                & ResNet-18-BN  & \textit{TCT}   &  \textbf{49.92\%}  & \textbf{83.02\%} & \textbf{89.21\%}  & \textbf{91.10\%}  \\
            \midrule
            \midrule
			\multirow{7}{*}{CIFAR-100}   &  &    &
			$\alpha=0.001$ & $\alpha=0.01$ & $\alpha=0.1$ & $\alpha=0.5$ 
			\\ 
			\cmidrule(lr){1-7} 
			  & ResNet-18-GN  & {FedAvg}  &   47.60\%  &  48.60\%  &  53.29\%  &   55.39\%    \\ 
			\tablewhitespace
			  & ResNet-18-BN & {FedAvg} & 53.89\% & 54.22\%  & 63.49\%  & 67.65\%    \\ 
			  \tablewhitespace
                & ResNet-18-BN  & \textit{TCT}  & \textbf{68.42\%}  & \textbf{69.07\%}  & \textbf{69.66\%}  & \textbf{69.68\%}   \\
			\bottomrule
		\end{tabular}
	\vspace{-.75em}
\end{table*}

\clearpage
\subsection{Additional Experimental Results of the Effect of Stage 1 Communication Round for TCT}
\begin{figure*}[ht]
\subcapcentertrue
  \begin{center}
    \subfigure[CIFAR10-(\texttt{\#C=2}), Train Accuracy.]{\includegraphics[width=0.49\textwidth]{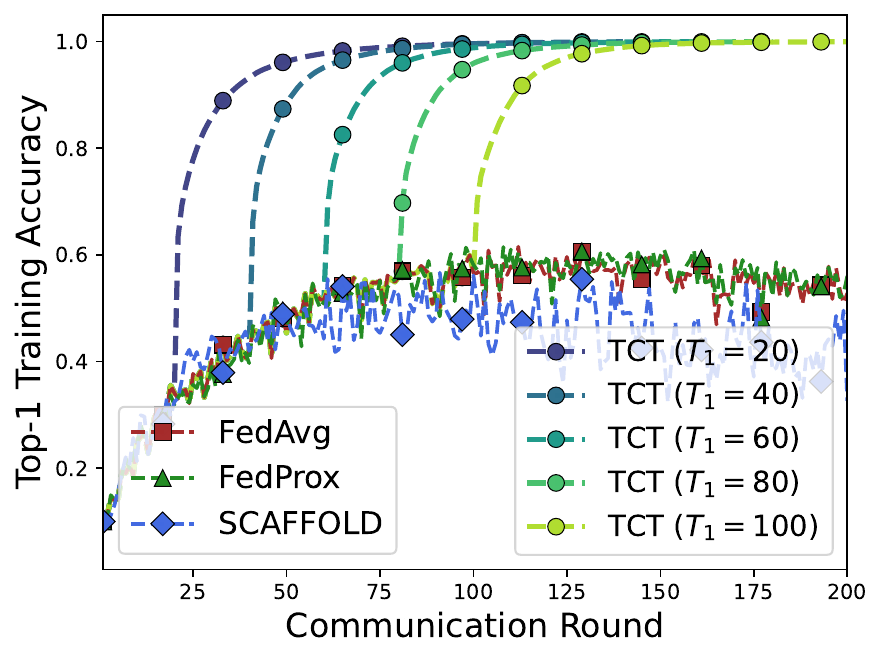}}
    \subfigure[CIFAR10-(\texttt{\#C=2}), Test Accuracy.]{\includegraphics[width=0.49\textwidth]{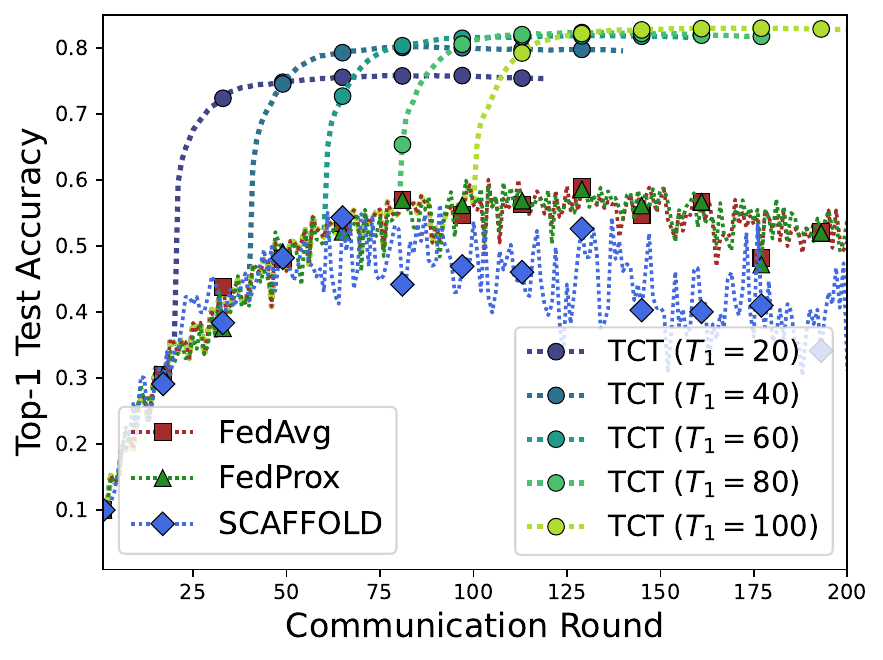}}
    \subfigure[CIFAR100-($\alpha=0.01$), Train Accuracy.]{\includegraphics[width=0.49\textwidth]{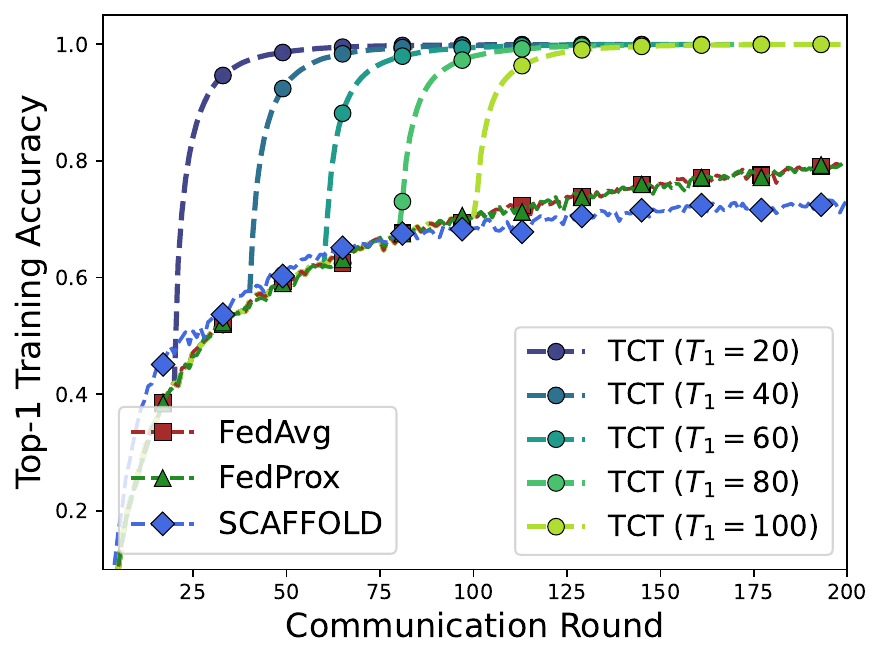}}
    \subfigure[CIFAR100-($\alpha=0.01$), Test Accuracy.]{\includegraphics[width=0.49\textwidth]{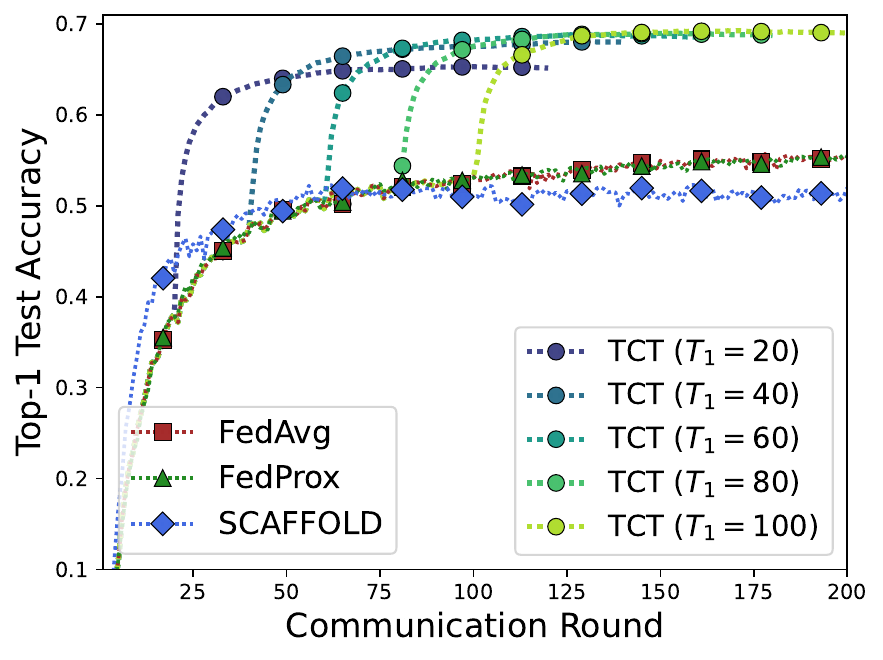}}
    \caption{ We evaluate TCT on using checkpoints saved at different communication rounds $T_1$ in \textbf{Stage 1}. We compare TCT to existing algorithm, including FedAvg, FedProx, and SCAFFOLD. For all three existing algorithms, we visualize the results of local learning rate $\eta=0.1$. 
    The train/test accuracy results in the first $T_1$ communication rounds of TCT are the same as FedAvg. For example, ``{TCT ($T_1=20$)}''  corresponds to training the model with FedAvg for $T_1=20$ rounds in \textbf{Stage 1} and then running $100$ rounds of SCAFFOLD for solving the linear regression problem in  \textbf{Stage 2}. Plots \textbf{(a)} and \textbf{(c)} display training accuracy and  \textbf{(b)} and \textbf{(d)} display test accuracy. 
    }
    \label{fig:appendix-fig-effect-T1}
  \end{center}
  \vskip -0.15in
\end{figure*}

In Figure~\ref{fig:appendix-fig-effect-T1}, we provide additional results of the effect of $T_1$ for TCT on CIFAR10 and CIFAR100 datasets. We find that TCT outperforms existing algorithm across all $T_1$ communication rounds, where $T_1\geq 20$. 
Extending the number of rounds for the baseline algorithms to 200 rounds does not improve their performance. In contrast, running 60 rounds of bootstrapping using FedAvg followed by 40 rounds of TCT gives near-optimal performance across all settings.

\clearpage
\subsection{Additional Experimental Results of Pre-trained Models}\label{subsec:pretraining-experiments}
\begin{figure*}[ht]
\subcapcentertrue
  \begin{center}
    \subfigure[CIFAR10-(\texttt{\#C=2}).]{\includegraphics[width=0.49\textwidth]{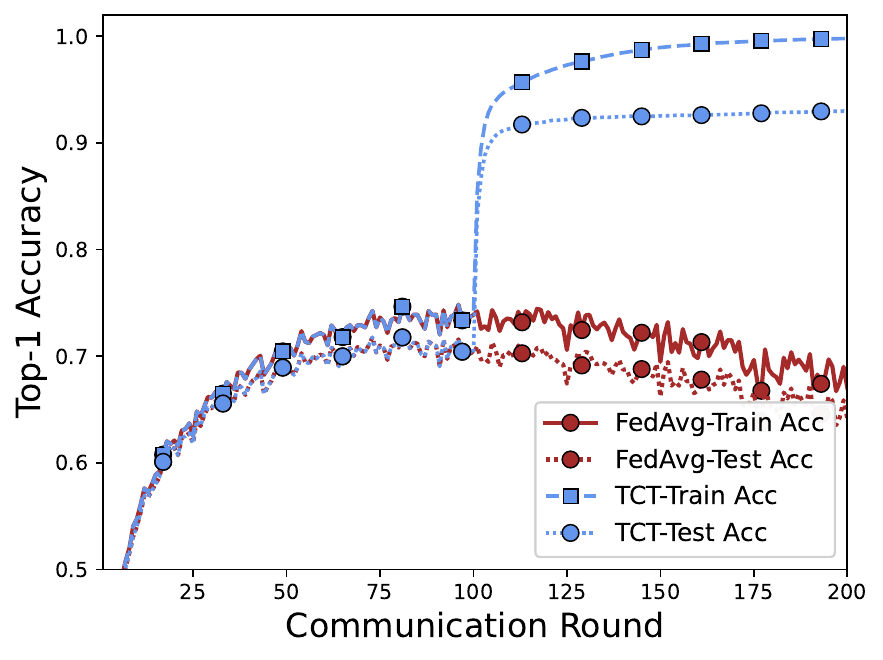}}
    \subfigure[CIFAR10-($\alpha=0.1$).]{\includegraphics[width=0.49\textwidth]{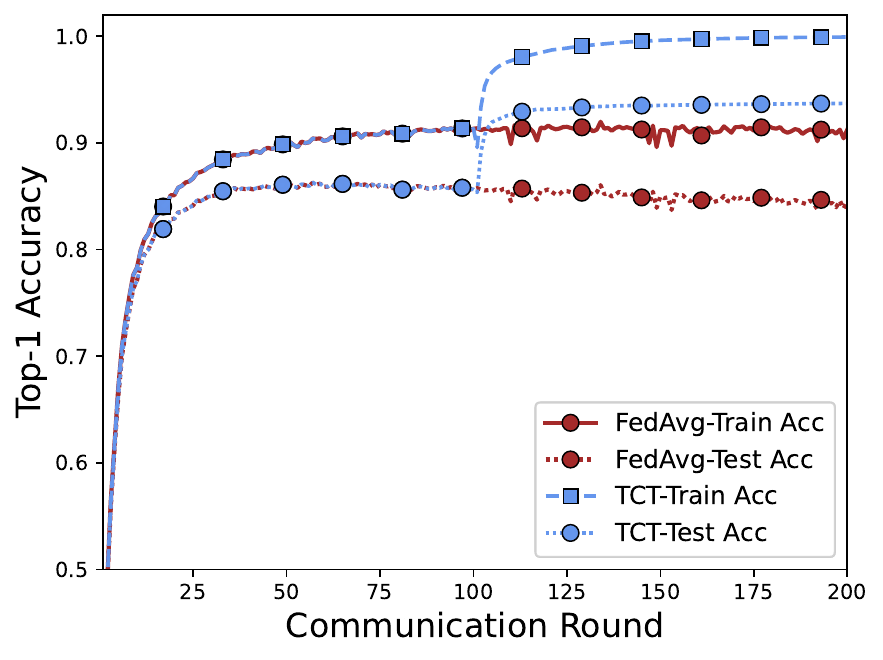}}
    \subfigure[CIFAR100-($\alpha=0.01$).]{\includegraphics[width=0.49\textwidth]{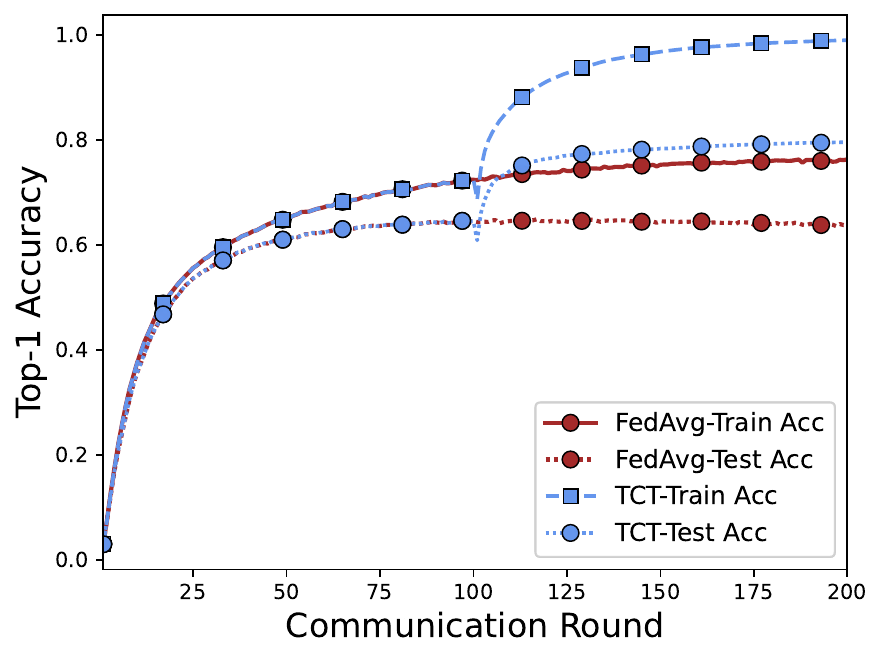}}
    \subfigure[CIFAR100-($\alpha=0.001$).]{\includegraphics[width=0.49\textwidth]{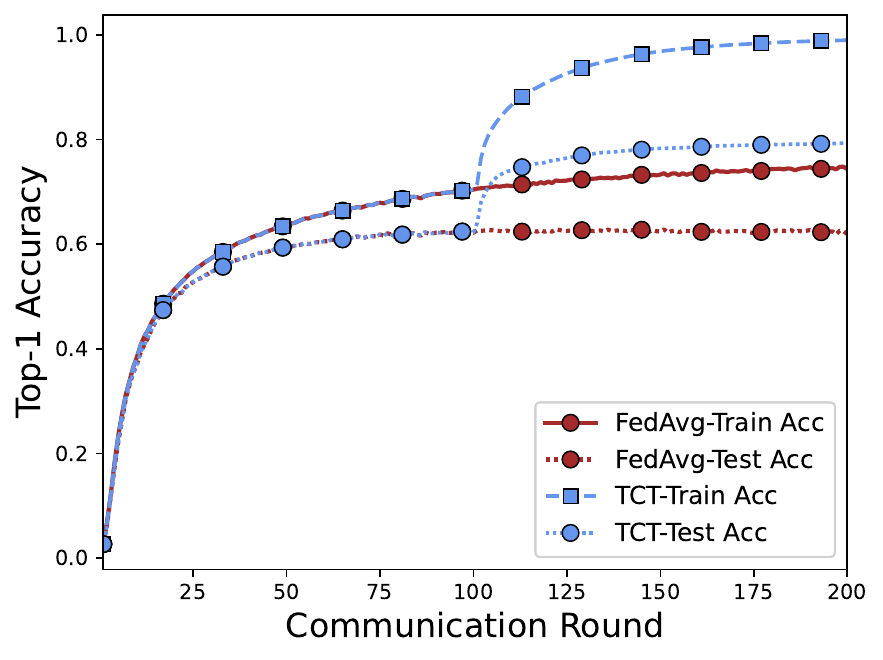}}
    \caption{ We evaluate FedAvg and TCT on CIFAR10 and CIFAR100 datasets with pre-trained ResNet-18. Plots \textbf{(a)} and \textbf{(b)} display training/test accuracy on the CIFAR10 dataset and  \textbf{(c)} and \textbf{(d)} display training/test accuracy on the CIFAR100 dataset. 
    }
    \label{fig:appendix-fig-effect-pretrain}
  \end{center}
  \vskip -0.15in
\end{figure*}

In Table~\ref{tab:table_pretrain} and  Figure~\ref{fig:appendix-fig-effect-pretrain}, we provide additional results of the effect of pre-training for FedAvg and TCT on CIFAR10 and CIFAR100 datasets. For both methods, we use the ResNet-18 pre-trained on ImageNet-1k~\citep{he2016deep} as the initialization. 
We use \textit{FedAvg (last layer)} to denote applying FedAvg on learning the last linear layer of the model, i.e., layers except for the last linear layer are freezed during training.
Compared to results in Table~\ref{tab:main_table}, we find that using pre-trained model as initialization largely improves the performance of both FedAvg and TCT. However, FedAvg still suffers from data heterogeneity. In contrast, TCT achieves similar performance as the centralized setting on both datasets across different degrees of data heterogeneity.

\begin{table*}[ht!]
	\centering
	\caption{{The top-1 test accuracy ($\%$) of our algorithm (TCT) vs.\ FedAvg evaluated on CIFAR10 and CIFAR100 with pre-trained model initialization. We vary the degree of data heterogeneity by controlling the $\alpha$ parameter of the symmetric Dirichlet distribution $\mathrm{Dir}_K(\alpha)$ and the $\texttt{\#C}$ parameter for assigning how many labels each client owns. Higher accuracy is better. The highest top-1 accuracy in each setting is highlighted in \textbf{bold}. } 
	}
	\vspace{0.75em}
	\label{tab:table_pretrain}
	\resizebox{0.99\textwidth}{!}{
	    \footnotesize
		\begin{tabular}{cccccc}
			\toprule
			\multirow{1}{*}{\bf Methods} & \multicolumn{5}{c}{\bf Datasets}
			  \\
			  \midrule
		\midrule
		   &   CIFAR10 ($\texttt{\#C}=2$) & CIFAR10 ($\alpha=0.1$)  & & CIFAR100 ($\alpha=0.001$) &  CIFAR100 ($\alpha=0.01$)   \\ 
			\cmidrule(lr){2-4}
			\cmidrule(lr){5-6}
			  {\color{gray}Centralized}  &  {\color{gray}95.13\%}   &  {\color{gray}95.13\%}  & & {\color{gray}80.65\%}  &   {\color{gray}80.65\%}   \\ 
			\tablewhitespace
			  FedAvg (last layer) & 63.60\% & 75.16\% & & 50.40\% &  51.97\%  \\ 
			\tablewhitespace
			  FedAvg  &  64.73\%   &  84.25\%  & & 62.23\%  &   63.81\%   \\ 
			\tablewhitespace
			  {TCT}  &   \textbf{92.97\%}   &  \textbf{93.70\%}  & & \textbf{79.25\%} &   \textbf{79.55\%}    \\ 
			\bottomrule
		\end{tabular}
	}
	\vspace{-.75em}
\end{table*}

\clearpage
\subsection{Additional Experimental Results of One-round Communication}
In Table~\ref{tab:table_oneround}, we provide additional results of TCT on CIFAR10 and CIFAR100 datasets with one communication round in TCT-Stage 2. Specifically, we set the number of local steps $M=500$, local learning rate $\eta=0.00005$, and the total number of communication round $T=1$ in TCT-Stage 2. The results are summarized in Table~\ref{tab:table_oneround}. With only one communication round in TCT-Stage 2, TCT still achieves better performance than FedAvg in three out of four settings in Table~\ref{tab:table_oneround}. On the other hand, we recommend setting the communication round in TCT-Stage 2 larger than 10 for our method TCT in order to achieve satisfying performance.

\begin{table*}[ht!]
	\centering
	\caption{{The top-1 test accuracy ($\%$) of our algorithm (TCT) on CIFAR10 and CIFAR100 with one communication round in TCT-Stage 2. } 
	}
	\vspace{0.75em}
	\label{tab:table_oneround}
	\resizebox{0.99\textwidth}{!}{
	    \footnotesize
		\begin{tabular}{cccccc}
			\toprule
			\multirow{1}{*}{\bf Methods} & \multicolumn{5}{c}{\bf Datasets}
			  \\
			  \midrule
		\midrule
		   &   CIFAR10 ($\texttt{\#C}=2$)   & CIFAR10 ($\alpha=0.01$) & & CIFAR100 ($\alpha=0.001$) & CIFAR100 ($\alpha=0.01$)   \\ 
			\cmidrule(lr){2-3}
			\cmidrule(lr){5-6}
			\tablewhitespace
			  FedAvg  &   56.86\%  & 82.60\% & &  53.89\% &     54.22\%   \\ 
			\tablewhitespace
			  {TCT}  &   83.02\%  &  89.21\% & &  68.42\% &     69.07\%   \\ 
			\tablewhitespace
			  {TCT-OneRound}  &  64.94\% &  82.62\% & & 55.50\%   & 57.51\%   \\ 
			\bottomrule
		\end{tabular}
	}
	\vspace{-.75em}
\end{table*}

\subsection{Additional Ablations}
\paragraph{Effect of local learning rate for TCT and FedAdam.} As mentioned in \citet{reddi2021adaptive}, FedAdam is more robust to the choice of local learning rate compared to FedAvg. We conduct additional ablations on the effect of local learning rate for TCT as well as FedAdam~\citep{reddi2021adaptive} on the CIFAR10 dataset. 
For each algorithm, we first select a base local learning rate $\eta_{\mathrm{base}}$ and then vary the local learning rate $\eta \in \{\eta_{\mathrm{base}}\cdot 10^{0}, \eta_{\mathrm{base}}\cdot 10^{-0.5},  \eta_{\mathrm{base}}\cdot 10^{-1.0}, \eta_{\mathrm{base}}\cdot 10^{-1.5}\}$. The results are summarized in Table~\ref{tab:compare_locallr_table}. Compared to FedAdam, we find that TCT is much less sensitive to the choice of local learning rate. 

\begin{table*}[ht]
	\centering
	\caption{{The top-1 test accuracy ($\%$) of our algorithm (TCT) and FedAdam~\citep{reddi2021adaptive} evaluated on  the  CIFAR10 dataset. We vary the local learning rate for both algorithms. Higher accuracy is better. } 
	}
	\vspace{0.75em}
	\label{tab:compare_locallr_table}
	\resizebox{0.97\textwidth}{!}{
	    \footnotesize
		\begin{tabular}{cccccc}
			\toprule
			\multirow{1}{*}{\bf Datasets} &   \multirow{1}{*}{\bf Methods}                 & \multicolumn{4}{c}{\bf Local learning rate}   
			  \\
            \midrule
		\midrule
		  &    ($\eta_{\mathrm{base}}=10^{-4}$)  & $\eta=\eta_{\mathrm{base}}\cdot 10^{0}$ & $\eta=\eta_{\mathrm{base}}\cdot 10^{-0.5}$   & $\eta=\eta_{\mathrm{base}}\cdot 10^{-1.0}$  & 	$\eta=\eta_{\mathrm{base}}\cdot 10^{-1.5}$ 
			\\ 
			\cmidrule(lr){1-6} 
			 CIFAR10-(\texttt{\#C=2}) &  TCT & 82.12\% & 83.60\% & 83.51\% & 82.37\% 
			  \\
			  \tablewhitespace 
			 CIFAR10-($\alpha=0.1$)  &  TCT & 88.68\% & 89.27\% & 89.23\% &  89.15\%
			 \\
            \midrule
            \midrule
			   &   ($\eta_{\mathrm{base}}=10^{-1.5}$)  & $\eta=\eta_{\mathrm{base}}\cdot 10^{0}$ & $\eta=\eta_{\mathrm{base}}\cdot 10^{-0.5}$   & $\eta=\eta_{\mathrm{base}}\cdot 10^{-1.0}$  & 	$\eta=\eta_{\mathrm{base}}\cdot 10^{-1.5}$ 
			\\ 
			\cmidrule(lr){1-6} 
			 CIFAR10-(\texttt{\#C=2})  & FedAdam~\citep{reddi2021adaptive} & 31.29\% & 33.52\% & 26.20\% & 14.96\%   \\ 
			\tablewhitespace
			 CIFAR10-($\alpha=0.1$) & FedAdam~\citep{reddi2021adaptive}  & 10.31\% & 37.26\% & 62.57\% & 49.18\%  \\ 
			\bottomrule
		\end{tabular}
	}
	\vspace{-.75em}
\end{table*}

\paragraph{Effect of local learning rate and number of local steps for TCT.} We conduct additional ablations on the effect of both the local learning rate $\eta$ and the number of local steps $M$ for TCT on CIFAR10 and CIFAR100 datasets.  The results in Table~\ref{tab:compare_locallr_M_table} and Table~\ref{tab:compare_locallr_M_table_CIFAR100} indicate that TCT is robust to the choice of the local learning rate $\eta$ and the number of local steps $M$. We find that as the number of steps increases, the learning rate should predictably decrease. The performance is relatively stable along the diagonal, indicating that it is the product $M \cdot \eta$ which affects accuracy.

\begin{table*}[ht]
	\centering
	\caption{{The top-1 test accuracy ($\%$) of our algorithm (TCT) evaluated on the CIFAR10 dataset. We consider CIFAR10-(\texttt{\#C=2}) and we set $\eta_{\mathrm{base}}=10^{-4}$. We vary both the local learning rate and the number of local steps for TCT.  Higher accuracy is better. } 
	}
	\vspace{0.75em}
	\label{tab:compare_locallr_M_table}
	\resizebox{0.97\textwidth}{!}{
	    \footnotesize
		\begin{tabular}{ccccc}
			\toprule
			\multirow{1}{*}{\bf Number of local steps}                 & \multicolumn{4}{c}{\bf Local learning rate}   
			  \\
            \midrule
		\midrule
		  &    $\eta=\eta_{\mathrm{base}}\cdot 10^{0}$ & $\eta=\eta_{\mathrm{base}}\cdot 10^{-0.5}$   & $\eta=\eta_{\mathrm{base}}\cdot 10^{-1.0}$  & 	$\eta=\eta_{\mathrm{base}}\cdot 10^{-1.5}$ 
			\\ 
			\cmidrule(lr){1-5} 
			 $M=50$ & 83.51\% & 82.37\% & 80.67\% & 78.14\% 
			  \\
			  \tablewhitespace 
			 $M=100$ & 83.59\% & 83.35\% & 81.73\% & 79.71\% 
			  \\
			  \tablewhitespace 
			 $M=500$ & 82.12\% & 83.60\% & 83.51\% & 82.37\% 
			  \\
			  \tablewhitespace 
			 $M=1000$ & 80.78\% & 82.92\% & 83.59\% & 83.35\% 
			  \\
			\bottomrule
		\end{tabular}
	}
	\vspace{-.75em}
\end{table*}

\begin{table*}[ht]
	\centering
	\caption{{The top-1 test accuracy ($\%$) of our algorithm (TCT) evaluated on the CIFAR100 dataset. We consider CIFAR100-($\alpha=0.01$) and we set $\eta_{\mathrm{base}}=10^{-4}$. We vary both the local learning rate and the number of local steps for TCT.  Higher accuracy is better. } 
	}
	\vspace{0.75em}
	\label{tab:compare_locallr_M_table_CIFAR100}
	\resizebox{0.97\textwidth}{!}{
	    \footnotesize
		\begin{tabular}{ccccc}
			\toprule
			\multirow{1}{*}{\bf Number of local steps}                 & \multicolumn{4}{c}{\bf Local learning rate}   
			  \\
            \midrule
		\midrule
		  &    $\eta=\eta_{\mathrm{base}}\cdot 10^{0}$ & $\eta=\eta_{\mathrm{base}}\cdot 10^{-0.5}$   & $\eta=\eta_{\mathrm{base}}\cdot 10^{-1.0}$  & 	$\eta=\eta_{\mathrm{base}}\cdot 10^{-1.5}$ 
			\\ 
			\cmidrule(lr){1-5} 
			 $M=50$ &  69.12\% &  67.31\% &  64.61\% & 	61.34\% 
			  \\
			  \tablewhitespace 
			 $M=100$ & 69.54\% &  68.48\% &  66.43\% &  63.42\%  
			  \\
			  \tablewhitespace 
			 $M=500$ & 69.03\% &  69.60\% &  69.12\% &  67.31\%  
			  \\
			  \tablewhitespace 
			 $M=1000$ & 68.38\% &  69.42\% &  69.54\% &  68.48\% 
			  \\
			\bottomrule
		\end{tabular}
	}
	\vspace{-.75em}
\end{table*}

\end{document}